\documentclass[aps,prl,twocolumn,superscriptaddress]{revtex4-2}

\usepackage{graphicx}

\usepackage{amssymb}
\usepackage{amsmath}
\usepackage{verbatim}

\usepackage{algorithmicx}
\usepackage{algpseudocode}

\begin{document}

\title{Infrared Universality of Collective Dynamics across Transformer and State-Space Architectures}

\author{Byung Gyu Chae}

\affiliation{Electronics and Telecommunications Research Institute, 218 Gajeong-ro, Yuseong-gu, Daejeon 34129, Republic of Korea
\\ bgchae@etri.re.kr}


\begin{abstract}

Whether distinct neural architectures develop common collective
dynamics remains an open question.
Recent analysis of Transformer language models revealed a nearly flat,
weakly infrared-enhanced time-scale density of states (TDOS) associated
with near-marginal long-memory dynamics.
Here we test whether a closely related organization emerges in Mamba,
whose selective state-space dynamics provides a fundamentally different
microscopic mechanism.

Mamba allows relaxation dynamics to be resolved at three levels: the
intrinsic spectrum of the learned state-space generator, its
input-conditioned selective rescaling, and the collective TDOS of the
complete block measured from its Jacobian.
These spectra are not identical: selective dynamics and the remaining
block transformations substantially reorganize the microscopic
relaxation hierarchy.
Nevertheless, the full block develops a reproducible slow-mode
continuum whose infrared sector becomes progressively better resolved
with increasing sequence length.

Cumulative spectral analysis over a fixed infrared window yields
$\rho(\lambda)\sim\lambda^\beta$, with the long-sequence Mamba
exponent stabilizing near $\beta_{\rm M}\simeq-0.17$.
The corresponding memory dynamics follows
$K(t)\sim t^{-(1+\beta)}$, close to the marginal $1/t$ regime, and is
confirmed directly from the measured relaxation spectrum.
Despite its different microscopic dynamics, Transformer exhibits a
closely related full-block infrared organization with representative
exponents of order $\beta_{\rm Tr}\sim-0.1$.

These results separate explicit microscopic state-space memory from
collective infrared organization and show that distinct sequence
architectures can develop closely related near-marginal slow-mode
dynamics.
They extend infrared collective organization beyond Transformers and
provide an independent test of the dynamical structure described by
Cognitive Field Theory.

\end{abstract}

\maketitle

\section{I. Introduction}

Large neural networks can exhibit collective dynamical behavior that
is not readily inferred from their microscopic computational
operations.
Although modern sequence models are built from well-defined
architectural components, their high-dimensional learned dynamics may
develop macroscopic organization that is less dependent on the
specific mechanisms used to process information.
Identifying such organization is important for determining whether
common dynamical principles can emerge across distinct neural
architectures.

Recent analysis of Transformer language models revealed a broad
time-scale density of states (TDOS) in the relaxation spectrum of
hidden-state dynamics, with substantial spectral weight extending
toward slow modes~\cite{1}.
The resolved infrared spectrum is nearly flat but weakly enhanced
toward small relaxation rates, approximately described by
$\rho(\lambda)\sim\lambda^\beta$ with $\beta\lesssim0$ and
$|\beta|\ll1$.
Such a spectrum corresponds to scale-free long-memory dynamics close
to the marginal $K(t)\sim1/t$ regime.
The persistence of this organization across training, input prompts,
network depth, and model scale suggests that it reflects a collective
property of Transformer dynamics rather than a feature of a particular
hidden-state realization.

A fundamental question is whether this infrared organization is
specific to Transformer computation.
Transformers share a common microscopic architecture based on
self-attention, feed-forward transformations, and residual
propagation~\cite{2,3,4,5,6}.
A stronger test therefore requires a sequence architecture whose
temporal dynamics is generated through a qualitatively different
mechanism.

State-space sequence models provide a distinct approach to long-range
temporal computation.
Modern developments progressed from continuous-time memory
representations such as HiPPO to structured and diagonal state-space
models designed for efficient long-sequence processing
~\cite{7,8,9,10}.
Subsequent work extended these ideas to language modeling and
attention-free long-range sequence processing, while related recurrent
formulations emphasized the importance of stable spectral
parameterization for long-time propagation~\cite{11,12,13}.
These developments established state-space dynamics as a major
alternative framework for representing long-range temporal structure
in learned sequence models.

Mamba provides a particularly important development along this
direction by introducing input-dependent selective state-space
dynamics~\cite{14}.
Rather than relying on self-attention, Mamba employs an internal
dynamical state whose evolution is controlled by input-dependent
state-space parameters.
More importantly for the present study, its construction exposes
several distinct levels of relaxation dynamics.
The learned state-space generator provides an intrinsic hierarchy of
relaxation scales, input-dependent selection temporally rescales this
hierarchy during inference, and the complete nonlinear block produces
a collective response that can be measured independently through its
Jacobian.
Mamba therefore provides a direct setting in which microscopic
state-space memory and macroscopic collective relaxation can be
distinguished within the same trained architecture.

The eigenvalue structure and parameterization of state-space
generators have been studied extensively in connection with stability,
memory, and long-range propagation~\cite{8,9,10,12}.
In selective SSMs, these intrinsic dynamics are further modulated by
input-dependent state transitions, providing a mechanism for
context-dependent retention and forgetting~\cite{14,15}.
A broad distribution of intrinsic SSM relaxation times does not by
itself establish collective infrared memory.
The standard SSM relaxation spectrum governs the propagation of input
history through the internal state, whereas collective infrared
organization concerns the relaxation structure of the complete learned
hidden-state transformation.
The central question is therefore whether the slow-mode hierarchy
encoded and selectively rescaled within the SSM survives or is
reorganized into a reproducible collective infrared spectrum at the
full-block level.

Here we examine this hierarchy in pretrained Mamba models.
We first characterize the intrinsic SSM relaxation spectrum and its
input-conditioned effective form, and then independently measure the
collective TDOS of the complete block.
These dynamical spectra are not identical: selective dynamics and the
remaining nonlinear block transformations substantially reorganize
the microscopic relaxation hierarchy.
Broad slow-mode structure is nevertheless observed at each level,
while the complete block develops a distinct collective spectrum with
a well-resolved infrared sector.

The collective measurements reveal a reproducible infrared scaling
regime.
As the retained sequence length is increased, the full-block TDOS
becomes progressively smoother and better resolved toward slow
relaxation rates, while its large-scale infrared organization remains
stable.
Cumulative spectral analysis over a fixed resolved infrared window
yields a weakly negative exponent that stabilizes near
$\beta_{\rm M}\simeq-0.17$ for the longest sequences examined.
The resulting spectrum is therefore nearly flat but weakly
infrared enhanced, corresponding through
$K(t)\sim t^{-(1+\beta)}$ to long-memory dynamics close to the
marginal $1/t$ form.

The central result emerges from comparison with Transformer dynamics.
Despite the fundamentally different microscopic mechanisms of
attention-based Transformers and selective state-space
models~\cite{2,14}, their full-block Jacobian spectra exhibit closely
related infrared organization.
Transformer measurements yield representative resolved exponents of
order $\beta_{\rm Tr}\sim-0.1$~\cite{1}, whereas the Mamba measurements
stabilize near $\beta_{\rm M}\sim-0.17$.
These values are neither identical nor interpreted here as a universal
critical exponent.
Rather, both architectures occupy a closely related near-marginal
regime characterized by a broad, nearly flat, weakly
infrared-enhanced collective TDOS.

Formal connections between attention and state-space models have also
been established through structured state-space duality~\cite{16}.
The comparison considered here is complementary: rather than relating
their computational operators, we compare the empirically measured
collective relaxation spectra of the complete learned blocks.

This comparison separates microscopic memory implementation from
collective dynamical organization.
Mamba implements temporal persistence explicitly through selective
state-space evolution, whereas Transformer blocks contain no
corresponding recurrent SSM state.
The appearance of closely related collective infrared spectra in both
systems therefore cannot be explained solely by the explicit
state-space memory mechanism of Mamba.
Instead, substantially different microscopic computations can produce
similar long-time organization at the level of the complete
hidden-state dynamics.

This observation is naturally connected with Cognitive Field Theory
~\cite{17}, in which long-lived macroscopic memory is associated with
the infrared organization of collective relaxation modes.
Within this framework, the TDOS determines the long-time memory kernel
through its Laplace transform, so that enhanced spectral weight at
slow relaxation rates generates long-lived non-Markovian response.
Mamba provides a particularly transparent test of this picture because
its microscopic state-space relaxation hierarchy can be examined
separately from the collective TDOS of the complete learned block.

The present results therefore extend the observation of near-marginal
infrared dynamics beyond the Transformer architecture.
They show that two substantially different sequence architectures can
develop closely related collective slow-mode organization despite
implementing temporal computation through different microscopic
mechanisms.
This separation between architecture-specific computation and
collective infrared organization suggests that near-marginal
slow-mode dynamics may represent a more general dynamical principle
of learned sequence processing.

The remainder of this paper is organized as follows.
Section~II introduces the hierarchy of relaxation dynamics in Mamba
and distinguishes the intrinsic and effective SSM spectra from the
full-block collective TDOS.
Section~III presents the spectral measurements and analyzes the
emergence and infrared scaling of the full-block dynamics.
Section~IV compares the collective spectra of Mamba and Transformer.
Section~V discusses the physical interpretation, limitations, and
implications of these results, followed by the conclusion in
Sec.~VI.

\section{II. Hierarchy of Relaxation Dynamics in Mamba}

Mamba provides a particularly useful architecture for investigating
the relation between microscopic relaxation dynamics and macroscopic
collective organization.
Unlike attention-based Transformers, where the relaxation spectrum is
reconstructed primarily from the Jacobian of the complete hidden-state
mapping, the state-space structure of Mamba exposes an internal
dynamical operator whose relaxation scales can be examined directly.
The selective mechanism further makes the effective state-space
evolution input dependent, while the complete Mamba block contains
additional projections, nonlinearities, gating operations, and
residual propagation.

The relaxation dynamics can therefore be examined at three distinct
levels,
\begin{equation}
A
\;\longrightarrow\;
\overline{A}_t(x)
\;\longrightarrow\;
J_{\rm block},
\label{eq:mamba_relaxation_hierarchy}
\end{equation}
corresponding respectively to the intrinsic state-space spectrum, the
input-conditioned effective SSM spectrum, and the collective response
of the complete Mamba block.
These objects are related through the architecture but describe
different levels of dynamical organization.
Distinguishing them is essential for determining whether the
infrared organization of the complete network simply reflects
relaxation scales already encoded in the microscopic SSM or emerges
through their selective and collective reorganization.

\subsection{A. Intrinsic state-space relaxation spectrum}

The dynamical core of a Mamba block is a continuous-time state-space
system of the form
\begin{align}
\frac{d h(t)}{dt}
&=
A h(t)
+
B x(t),
\\
y(t)
&=
C h(t),
\label{eq:ssm_continuous}
\end{align}
where $x(t)$ denotes the input, $h(t)$ the internal state, and
$A$ the state-transition generator.
The eigenvalue spectrum of $A$ defines the intrinsic dynamical
time scales available to the state-space system before
input-dependent selective modulation is applied.

For an eigenvalue
$a_i\in{\rm eig}(A)$, a stable continuous-time mode satisfies
${\rm Re}\,a_i<0$, and its intrinsic relaxation rate is
\begin{equation}
\lambda_i^{\rm int}
=
-{\rm Re}\,a_i.
\label{eq:intrinsic_relaxation_rate}
\end{equation}
The corresponding relaxation time is
$\tau_i^{\rm int}=1/\lambda_i^{\rm int}$.
In the standard Mamba parameterization used below,
$A=-\exp(A_{\log})$, so that the intrinsic relaxation rates are
directly given by $\lambda_A=\exp(A_{\log})$.

The intrinsic relaxation spectrum therefore characterizes the
hierarchy of time scales encoded directly in the learned SSM
generator.
We define its normalized relaxation spectral density as
\begin{equation}
\rho_{\rm int}(\lambda)
=
\frac{1}{N_{\rm int}}
\sum_i
\delta
\left(
\lambda-\lambda_i^{\rm int}
\right),
\label{eq:intrinsic_spectral_density}
\end{equation}
where $N_{\rm int}$ denotes the number of intrinsic state-space
modes included in the measurement.

An important property of this spectrum is that it can be measured
without specifying an input sequence.
It therefore provides a direct probe of the relaxation structure
encoded in the learned state-space parameters themselves.
In this sense, $\rho_{\rm int}(\lambda)$ represents the microscopic
dynamical spectrum from which the input-conditioned SSM dynamics is
constructed.

This intrinsic spectrum should not, however, be identified with the
collective TDOS of the complete Mamba block.
The state transition used during sequence processing is modified by
the selective mechanism, while the complete block contains additional
transformations outside the SSM.
The intrinsic spectrum therefore establishes the microscopic
relaxation structure against which the subsequent input-conditioned
and collective levels can be compared.

\subsection{B. Input-conditioned effective SSM dynamics}

During sequence processing, Mamba does not propagate the continuous
state directly through the bare generator $A$.
Instead, the continuous-time dynamics is discretized using an
input-dependent step size.
Writing the selective discretization parameter at token position $t$
as $\Delta_t(x)$, the effective state-transition operator is
schematically
\begin{equation}
\overline{A}_t(x)
=
\exp
\left[
\Delta_t(x) A
\right].
\label{eq:effective_A}
\end{equation}
The dependence of $\Delta_t$ on the processed input makes the
effective relaxation dynamics input and token dependent.

If
$\mu_{i,t}^{\rm eff}$ is an eigenvalue of
$\overline{A}_t(x)$, the corresponding discrete effective relaxation
rate is
\begin{equation}
\lambda_{i,t}^{\rm eff}
=
-\ln
\left|
\mu_{i,t}^{\rm eff}
\right|.
\label{eq:effective_relaxation_rate}
\end{equation}
For the exponential discretization above, a stable intrinsic mode
with $\lambda_i^{\rm int}=-{\rm Re}\,a_i$ therefore satisfies
\begin{equation}
\lambda_{i,t}^{\rm eff}
=
\Delta_t(x)\lambda_i^{\rm int}.
\label{eq:effective_intrinsic_relation}
\end{equation}

This relation makes the distinction between the intrinsic and
effective spectra explicit.
The intrinsic spectrum specifies the relaxation hierarchy encoded in
the learned SSM generator, whereas the selective mechanism locally
rescales these rates according to the processed input.
Consequently, even for a fixed learned matrix $A$, different token
positions and input sequences can realize different effective
relaxation structures.

To characterize this input-conditioned dynamics over a sequence, we
collect the effective relaxation rates into the normalized effective
SSM spectral density
\begin{equation}
\rho_{\rm eff}(\lambda)
=
\frac{1}{N_{\rm eff}}
\sum_{t,i}
\delta
\left(
\lambda-\lambda_{i,t}^{\rm eff}
\right),
\label{eq:effective_spectral_density}
\end{equation}
where $N_{\rm eff}$ is the total number of sampled effective
state-space modes.

This construction introduces an important distinction from the
intrinsic spectrum.
Increasing the retained sequence length does not modify the learned
matrix $A$, but samples a larger and more diverse set of
input-conditioned state transitions.
The sequence-length dependence of $\rho_{\rm eff}(\lambda)$ therefore
reveals how selective dynamics populates and redistributes the
effective relaxation scales encountered during sequence processing.

The effective SSM spectrum nevertheless remains an internal
state-space quantity.
It describes the input-conditioned propagation of the SSM state but
does not contain the complete response of the Mamba block.
Input and output projections, gating, nonlinear transformations,
channel mixing, and residual propagation can further reorganize the
dynamics seen by the hidden representation.
The collective response must therefore be characterized separately
through the Jacobian of the complete block.

\subsection{C. Full-block collective relaxation spectrum}

Let
\begin{equation}
H_{\ell+1}
=
F_\ell(H_\ell)
\label{eq:mamba_block_map}
\end{equation}
denote the complete hidden-state mapping implemented by Mamba block
$\ell$, where $H_\ell$ contains the retained sequence
representations.
The local response of the complete block around a given hidden state
is described by the Jacobian
\begin{equation}
J_\ell
=
\frac{
\partial\,{\rm vec}(H_{\ell+1})
}{
\partial\,{\rm vec}(H_\ell)
}.
\label{eq:mamba_full_jacobian}
\end{equation}

Unlike the intrinsic and effective SSM operators, $J_\ell$ contains
the response of the complete block.
Its spectrum therefore incorporates not only the state-space dynamics
and input-dependent selection, but also nonlinear transformations,
gating, channel mixing, output projection, and residual propagation.

For a Jacobian eigenvalue
\begin{equation}
\zeta_\alpha
=
|\zeta_\alpha|e^{i\theta_\alpha},
\end{equation}
we define the collective relaxation rate as
\begin{equation}
\lambda_\alpha^{\rm block}
=
-\ln|\zeta_\alpha|.
\label{eq:block_relaxation_rate}
\end{equation}
Modes with $0<|\zeta_\alpha|<1$ have positive relaxation rates and
correspond to locally contracting directions.
As $|\zeta_\alpha|$ approaches unity from below,
$\lambda_\alpha^{\rm block}$ approaches zero, corresponding to
increasingly slow collective relaxation.

The corresponding collective time-scale density of states (TDOS) is
defined as
\begin{equation}
\rho_{\rm block}(\lambda)
=
\frac{1}{N_{\rm block}}
\sum_\alpha
\delta
\left(
\lambda-\lambda_\alpha^{\rm block}
\right).
\label{eq:block_tdos}
\end{equation}
Unlike the intrinsic and effective SSM spectral densities,
$\rho_{\rm block}(\lambda)$ characterizes the collective relaxation
spectrum of the complete hidden-state transformation.
It is also the direct analogue of the collective TDOS measured from
Transformer block Jacobians and therefore provides a common
observable for comparing the infrared organization of the two
architectures.

The three spectra introduced above consequently describe distinct
levels of Mamba dynamics.
The intrinsic spectral density $\rho_{\rm int}(\lambda)$ characterizes
relaxation scales encoded in the learned SSM generator,
$\rho_{\rm eff}(\lambda)$ describes their input-conditioned selective
rescaling, and the collective TDOS $\rho_{\rm block}(\lambda)$
characterizes the response of the complete Mamba block.
They should therefore not be interpreted as different
representations of the same spectral object.

There is likewise no requirement that the three spectra have the same
functional form.
Differences among the intrinsic, effective, and full-block spectra
instead reveal how the internal state-space relaxation hierarchy is
progressively reorganized by selective dynamics and by the remaining
transformations of the complete block.
The measurements below use this hierarchy to determine which spectral
features are already present microscopically and which persist or
emerge at the collective level.

\subsection{D. From microscopic relaxation to macroscopic infrared organization}

The hierarchy introduced above makes it possible to distinguish
microscopic state-space memory from macroscopic collective relaxation
within the same Mamba architecture.
This distinction is important because the convolution kernel of a
standard state-space model and the memory kernel appearing in a
self-consistent collective response theory have different dynamical
roles, even though both can be constructed from distributions of
relaxation modes.

For a continuous-time linear state-space system,
\[
\dot h(t)
=
A h(t)+B x(t),
\,\,\,
y(t)
=
C h(t),
\label{eq:ssm_linear_system}
\]
the formal solution is
\begin{equation}
h(t)
=
e^{At}h(0)
+
\int_0^t
d\tau\,
e^{A(t-\tau)}
B x(\tau),
\label{eq:ssm_formal_solution}
\end{equation}
so that
\begin{equation}
y(t)
=
C e^{At}h(0)
+
\int_0^t
d\tau\,
K_{\rm SSM}(t-\tau)x(\tau),
\label{eq:ssm_convolution}
\end{equation}
with the standard SSM convolution kernel
\begin{equation}
K_{\rm SSM}(t)
=
C e^{At}B.
\label{eq:ssm_kernel}
\end{equation}

If the state generator is decomposed into relaxation modes,
$A v_\alpha=-\lambda_\alpha v_\alpha$, the kernel can be written
spectrally as
\begin{equation}
K_{\rm SSM}(t)
=
\sum_\alpha
c_\alpha e^{-\lambda_\alpha t},
\label{eq:ssm_mode_kernel}
\end{equation}
where the coefficients $c_\alpha$ depend on the input and output
couplings.
The corresponding frequency-domain transfer function is
\begin{equation}
G_{\rm SSM}(\omega)
=
C(-i\omega I-A)^{-1}B.
\label{eq:ssm_transfer_function}
\end{equation}

The standard SSM kernel therefore propagates and filters the history
of the external input.
Its memory is an input-history memory generated by the internal
state-space dynamics.
The existence of a broad intrinsic relaxation spectrum by itself does
not imply a self-energy dressing of a macroscopic collective field.
A broad distribution of relaxation times is therefore not, by itself,
evidence for collective infrared memory: the dynamical question is
whether these modes merely transmit past input or participate in a
closed response structure that reorganizes the collective dynamics
itself.

This distinction becomes explicit in a self-consistent
relaxation-field construction.
Consider a collective field $\phi(t)$ coupled to internal relaxation
modes $q_\alpha(t)$,
\begin{align}
\dot\phi(t)
&=
-r\phi(t)
+
\sum_\alpha g_\alpha q_\alpha(t)
+
\eta(t),
\\
\dot q_\alpha(t)
&=
-\lambda_\alpha q_\alpha(t)
+
g_\alpha\phi(t).
\label{eq:coupled_relaxation_field}
\end{align}
Solving for the relaxation modes gives
\begin{equation}
q_\alpha(t)
=
q_\alpha(0)e^{-\lambda_\alpha t}
+
g_\alpha
\int_0^t
d\tau\,
e^{-\lambda_\alpha(t-\tau)}
\phi(\tau).
\label{eq:qalpha_solution}
\end{equation}
Substitution back into the collective-field equation yields
\begin{equation}
\dot\phi(t)
=
-r\phi(t)
+
\int_0^t
d\tau\,
K_{\rm coll}(t-\tau)\phi(\tau)
+
\eta_{\rm eff}(t),
\label{eq:collective_nonmarkovian}
\end{equation}
where
\begin{equation}
K_{\rm coll}(t)
=
\sum_\alpha
g_\alpha^2 e^{-\lambda_\alpha t}.
\label{eq:collective_memory_kernel}
\end{equation}

The essential structural distinction is therefore between
$K_{\rm SSM}*x$ and $K_{\rm coll}*\phi$.
In the standard SSM, the relaxation kernel transmits the history of
an external input through the internal state.
In the collective equation, by contrast, the collective variable
excites the relaxation sector and subsequently receives its own
delayed response.
The latter therefore forms a closed dynamical feedback channel.

This difference becomes particularly transparent in frequency space.
Eliminating the relaxation modes gives
\begin{equation}
\left[
-i\omega+r
-
\Sigma_R(\omega)
\right]
\phi(\omega)
=
\eta_{\rm eff}(\omega),
\end{equation}
with
\begin{equation}
\Sigma_R(\omega)
=
\sum_\alpha
\frac{g_\alpha^2}
{\lambda_\alpha-i\omega}.
\label{eq:relaxation_self_energy}
\end{equation}
Defining $G_0^{-1}(\omega)=-i\omega+r$, the dressed collective
response therefore takes the Dyson form
\begin{equation}
G_R^{-1}(\omega)
=
G_0^{-1}(\omega)
-
\Sigma_R(\omega).
\label{eq:dyson_collective}
\end{equation}
The relaxation sector thus modifies the propagator of the collective
variable itself, rather than merely filtering an externally supplied
input history.
This self-energy structure is not generated by the standard
input-driven SSM convolution alone; it requires a closed response
channel through which the collective variable is dynamically coupled
back to the relaxation sector.

The relevance of this distinction for Mamba is immediate.
The intrinsic rates encoded in the learned SSM generator and their
input-conditioned selective rescaling characterize the explicitly
constructed state-space memory mechanism.
The complete Mamba block, however, combines this internal dynamics
with projections, gating, nonlinear transformations, channel mixing,
and residual propagation.
Its collective relaxation must therefore be characterized
independently from the Jacobian of the complete hidden-state
transformation.

Using the quantities defined above, the microscopic-to-collective
hierarchy examined in this work is
\begin{equation}
\lambda_A
\longrightarrow
\lambda_{\rm eff}(x_t)
\longrightarrow
\{\lambda_\alpha^{\rm block}\}
\longrightarrow
\rho_{\rm block}(\lambda).
\label{eq:mamba_micro_macro_hierarchy}
\end{equation}
The first two levels characterize the explicitly constructed
state-space relaxation mechanism, whereas
$\rho_{\rm block}(\lambda)$ is the collective TDOS of the complete
learned block.

This distinction provides the basis for the measurements below.
A broad intrinsic or effective SSM spectrum alone does not establish
collective infrared organization.
The relevant empirical question is whether, after selective
conditioning and the remaining nonlinear block transformations, the
independently measured full-block Jacobian develops a reproducible
low-relaxation-rate collective spectrum.
If that spectrum remains stable as the retained sequence subspace is
enlarged and persists across network depth, the infrared organization
cannot be reduced simply to the microscopic convolution spectrum of
the underlying SSM.

\begin{figure*}[t]
\centering
\includegraphics[width=1.0\textwidth, trim=0cm 6.5cm 0cm 0cm]{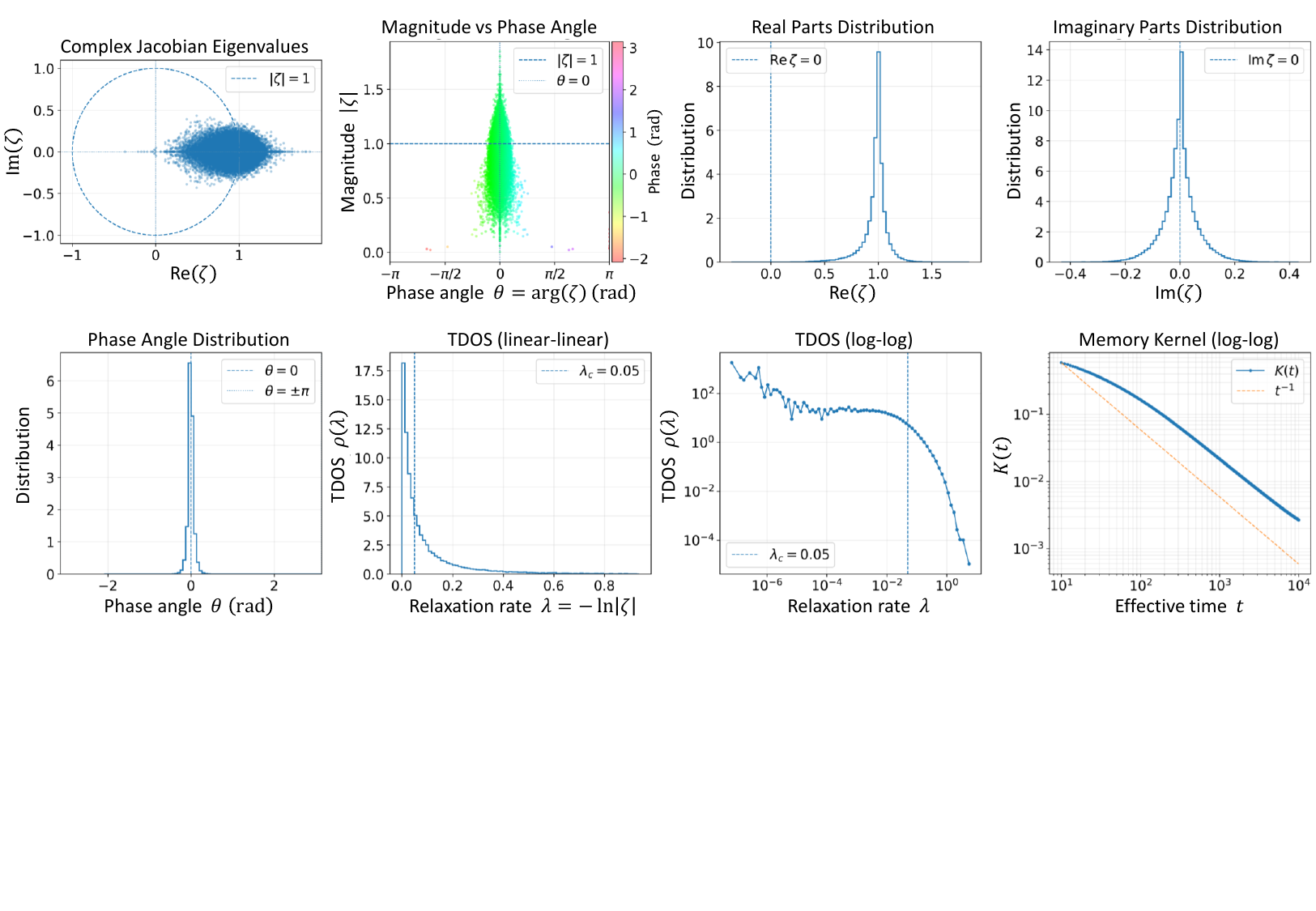}
\caption{
Collective spectral analysis of a representative Mamba-370M block
(block index \(\ell=24\), \(L=128\) retained tokens).
(a) Complex eigenvalue spectrum of the full-block Jacobian.
(b) Eigenvalue magnitude as a function of phase angle.
(c,d) Distributions of the real and imaginary parts of the complex
eigenvalues.
(e) Phase-angle distribution, strongly concentrated near
\(\theta=0\).
(f,g) Time-scale density of states (TDOS) obtained from the stable
eigenvalue magnitudes through
\(\lambda=-\ln|\zeta|\), shown on linear and logarithmic scales.
The measured spectrum spans many orders of magnitude in relaxation
rate and contains a broad population of slow collective modes.
The reference scale
\(\lambda_c=5\times10^{-2}\) is shown only as a visual guide to the
slow-relaxation sector and does not define a physical spectral
boundary.
(h) Memory kernel obtained directly from the Laplace transform of the
measured full-block TDOS, without assuming a particular spectral
functional form.
The quantitative sequence-length dependence, infrared convergence,
and scaling of the collective slow-mode sector are analyzed in the
following sections.
}
\label{fig:phase_portrait}
\end{figure*}

\section{III. Emergence of the Collective Infrared Spectrum}

Having distinguished the intrinsic, effective, and full-block
relaxation dynamics, we now examine how these levels are related in a
trained Mamba model.
We begin by defining the spectral measurement used to characterize
the collective dynamics of a complete Mamba block.

\subsection{A. Collective spectral measurement of a Mamba block}

Throughout this work, we analyze the publicly available pretrained
Mamba-370M base model.
The model contains 48 Mamba layers with hidden dimension
$d_{\rm model}=1024$ and was pretrained on 300 billion tokens from
the Pile.
The principal architectural parameters relevant to the present
spectral analysis are summarized in Table~\ref{tab:mamba370m}.

We focus on a single model scale in order to examine, within the same
learned architecture, the hierarchy from the intrinsic SSM
relaxation spectrum to the input-conditioned effective spectrum and
finally to the collective spectrum of the complete block.
The retained sequence length is varied systematically from
$L=8$ to $L=512$, allowing the resolved infrared structure to be
tested as increasingly large sequence subspaces are included.

\begin{table}[t]
\centering
\caption{
Configuration of the pretrained Mamba-370M model used throughout
this work. The SSM parameters correspond to the standard Mamba
configuration used by the released model.
}
\label{tab:mamba370m}
\begin{tabular*}{\columnwidth}{@{\extracolsep{\fill}}lc@{}}
\hline\hline
Configuration & Mamba-370M \\
\hline
Model parameters & 370M \\
Number of Mamba layers, $N_{\rm layer}$ & 48 \\
Hidden dimension, $d_{\rm model}$ & 1024 \\
SSM state dimension, $d_{\rm state}$ & 16 \\
Expansion factor & 2 \\
Local convolution width, $d_{\rm conv}$ & 4 \\
Pretraining tokens & 300B \\
Pretraining corpus & The Pile \\
\hline\hline
\end{tabular*}
\end{table}

For a hidden representation $H_\ell$ entering a Mamba block, we write
the complete block transformation as
\[
H_{\ell+1}
=
F_\ell(H_\ell).
\]
The collective relaxation spectrum is obtained from the Jacobian of
this complete hidden-state transformation,
\begin{equation}
J_{\rm block}
=
\frac{
\partial\,{\rm vec}(H_{\ell+1})
}{
\partial\,{\rm vec}(H_\ell)
}.
\label{eq:mamba_block_jacobian}
\end{equation}
Because this Jacobian contains the response of the entire block, its
eigenvalue spectrum provides a local measurement of the collective
dynamics seen by the hidden representation.

For a retained sequence of length $L$, the Jacobian can be decomposed
into token-resolved blocks
\begin{equation}
J_{ts}
=
\frac{\partial h'_t}{\partial h_s},
\qquad
t,s=1,\ldots,L.
\label{eq:mamba_token_jacobian}
\end{equation}
Since the Mamba sequence mapping is causal, the output at token
position $t$ cannot depend on future input positions, and therefore
\begin{equation}
J_{ts}=0,
\qquad
s>t.
\end{equation}
The full sequence Jacobian is consequently block lower triangular.
Its characteristic polynomial factorizes into those of the
token-diagonal blocks, implying the exact spectral identity
\begin{equation}
\sigma(J_{\rm block})
=
\bigcup_{t=1}^{L}
\sigma(J_{tt}).
\label{eq:mamba_causal_spectral_identity}
\end{equation}
Accordingly, the full eigenvalue spectrum can be obtained from the
token-diagonal Jacobians without explicitly constructing and
diagonalizing the much larger $(Ld_{\rm model})\times
(Ld_{\rm model})$ sequence Jacobian.
The derivation of Eq.~(\ref{eq:mamba_causal_spectral_identity}) and
the distinction between this eigenvalue reduction and the
off-diagonal propagation dynamics are given in Appendix~A.

Figure~1 illustrates the complete spectral analysis for a
representative Mamba-370M block.
The measurement is performed at layer~24 using a retained sequence
subspace of $L=128$ tokens.
Throughout this work, layer indices follow the zero-based block
indexing of the model implementation; thus, layer~24 denotes the
25th Mamba block in one-based counting.

Diagonalization of the token-resolved Jacobian blocks gives complex
eigenvalues
\[
\zeta_{t\alpha}
=
|\zeta_{t\alpha}|e^{i\theta_{t\alpha}},
\]
whose union, by Eq.~(\ref{eq:mamba_causal_spectral_identity}), is the
eigenvalue spectrum of the full causal sequence Jacobian.
For notational simplicity, we suppress the token index below and
write the pooled spectrum as $\{\zeta_\alpha\}$.

The eigenvalue magnitudes define the relaxation rates
\begin{equation}
\lambda_\alpha
=
-\ln|\zeta_\alpha|,
\label{eq:block_lambda_sec3}
\end{equation}
while their phases
$\theta_\alpha=\arg(\zeta_\alpha)$ characterize the oscillatory or
circulation sector of the local linearized dynamics.

Here propagation through one complete network block is taken as one
unit of discrete evolution, $\Delta s=1$, so that
$|\zeta_\alpha|=e^{-\lambda_\alpha\Delta s}$ directly gives
Eq.~(\ref{eq:block_lambda_sec3}).
The resulting $\lambda_\alpha$ is therefore a dimensionless
relaxation rate per block-propagation step rather than a rate per unit
of physical time.

The same convention was used in our Transformer analysis~\cite{1},
permitting direct comparison of the full-block relaxation spectra of
the two architectures.
This comparison does not apply to the absolute values of the
intrinsic or selectively rescaled SSM rates, which are defined in
Mamba's internal continuous-time parameterization.

Figures~1(a)--1(e) show complementary representations of the complex
collective spectrum.
The eigenvalues occupy a broad region of the complex plane, with a
large population concentrated near the positive real axis and the
phase distribution strongly centered around $\theta=0$.
The analysis below focuses on the relaxation sector obtained from the
eigenvalue magnitudes through Eq.~(\ref{eq:block_lambda_sec3}).

From the stable relaxation modes we construct the time-scale density
of states
\begin{equation}
\rho_L(\lambda)
=
\frac{1}{N}
\sum_{\alpha}
\delta(\lambda-\lambda_\alpha),
\label{eq:mamba_tdos_sec3}
\end{equation}
where $N$ denotes the total number of stable modes included in the
relaxation spectrum.

Equivalently, defining the empirical TDOS of token position $t$ as
$\rho_t(\lambda)$, the causal spectral decomposition implies
\begin{equation}
\rho_L(\lambda)
=
\frac{1}{L}
\sum_{t=1}^{L}
\rho_t(\lambda),
\label{eq:mamba_token_tdos_average}
\end{equation}
up to the corresponding stable-mode normalization when the number of
retained stable modes varies between token positions.
Thus the sequence-level TDOS is directly constructed from the
token-resolved relaxation spectra.

Equation~(\ref{eq:mamba_token_tdos_average}) also makes it possible to
distinguish the spectral structure itself from finite-sequence
sampling.
If the token-resolved spectra sample a common learned dynamical
organization, individual $\rho_t(\lambda)$ fluctuate around a common
spectral envelope, whereas increasing $L$ suppresses finite-mode
fluctuations and progressively improves the resolution of the same
underlying distribution.
The statistical self-averaging of the token-resolved spectra,
including the enhanced finite-sampling fluctuations expected in the
deep infrared, is developed in Appendix~B.

The resulting TDOS is shown in linear and logarithmic representations
in Figs.~1(f) and 1(g).
The measured spectrum covers many orders of magnitude in relaxation
rate, from a deeply resolved low-$\lambda$ sector to rates of order
unity and above.
A broad population of slow collective modes extends continuously
toward the smallest numerically resolved relaxation rates, while the
spectrum remains broadly populated throughout the intermediate
relaxation regime.
Thus, already at a single representative layer and sequence length,
the complete Mamba block exhibits a wide hierarchy of collective
relaxation time scales.

The precise location of the extreme low-$\lambda$ tail should not,
however, be interpreted as a physical infrared cutoff.
Modes in this regime are the most sensitive to finite spectral
resolution, finite sampling, and numerical precision.
We therefore base the quantitative infrared analysis below on a
common resolved fitting interval rather than on the smallest observed
relaxation rate.
Independent calculations using double-precision arithmetic confirm
that the broad slow-mode organization is robust.
The numerical behavior of the extreme infrared tail is examined
separately in the corresponding numerical-robustness appendix, while
its finite-sampling behavior is discussed analytically in
Appendix~C.

The dynamical consequence of the measured relaxation spectrum can be
represented by the corresponding memory kernel,
\begin{equation}
K(t)
=
\int d\lambda\,
\rho_L(\lambda)e^{-\lambda t}
=
\frac{1}{N}
\sum_{\alpha}
e^{-\lambda_\alpha t}.
\label{eq:mamba_memory_kernel}
\end{equation}
Thus the memory kernel is the Laplace transform of the measured TDOS.

Figure~1(h) shows the resulting broad temporal response, reflecting
the contribution of relaxation modes distributed over many time
scales.
Its quantitative long-time scaling is analyzed below together with
the infrared spectrum.

At this stage, the enhancement observed at very small $\lambda$
should not yet be interpreted as evidence for a particular infrared
power law.
Finite sequence length limits the number of resolved collective modes,
and the deepest infrared sector is consequently the most sensitive to
spectral sampling.
The sequence-length dependence, convergence of the collective
spectrum, and quantitative infrared exponent are therefore examined
in the following subsections.

Figure~1 establishes the more basic empirical observation that the
full-block dynamics contains a broad and densely populated
slow-relaxation sector.

The measurement procedure used throughout the remainder of the paper
can therefore be summarized as
\begin{equation}
J_{\rm block}
\rightarrow
\{J_{tt}\}
\rightarrow
\{\zeta_\alpha\}
\rightarrow
\{\lambda_\alpha,\theta_\alpha\}
\rightarrow
\rho_L(\lambda)
\rightarrow
K(t).
\label{eq:mamba_measurement_chain}
\end{equation}

Figure~1 defines this collective observable; Appendix~A establishes
the exact causal spectral reduction underlying its measurement,
Appendix~B develops its token-resolved self-averaging properties, and
the following analysis examines its microscopic origin,
sequence-length dependence, and infrared scaling.

\begin{figure*}[t]
\centering
\includegraphics[width=1.0\textwidth, trim=0cm 11.1cm 0cm 0cm]{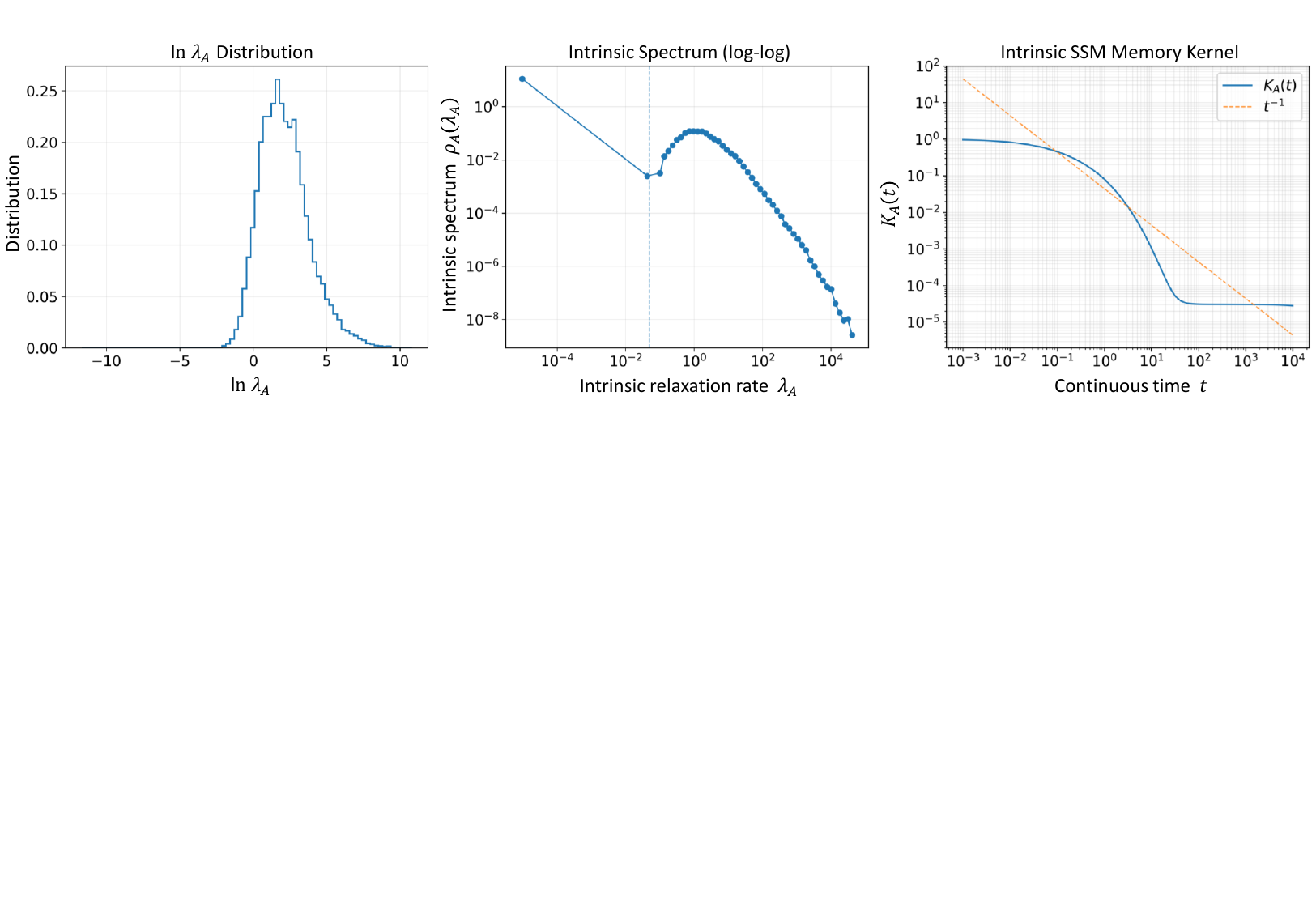}
\caption{
Intrinsic relaxation spectrum of the learned SSM in
Mamba-370M.
(a) Distribution of the learned logarithmic state-space parameters
$A_{\log}$ for a representative block.
The broad distribution is mapped to intrinsic continuous-time
relaxation rates through
$\lambda_A=\exp(A_{\log})$.
(b) Corresponding intrinsic relaxation-rate density
$\rho_A(\lambda_A)$ on logarithmic axes.
The spectrum extends over many orders of magnitude, from a
slow-relaxation branch near $\lambda_A\sim10^{-5}$ to rapidly
relaxing modes beyond $\lambda_A\sim10^{4}$, demonstrating that a
broad hierarchy of intrinsic time scales is already encoded in the
learned SSM generator.
(c) Intrinsic spectral relaxation kernel
$K_A(t)=\int d\lambda_A\,\rho_A(\lambda_A)e^{-\lambda_A t}$
computed directly from the measured intrinsic relaxation spectrum.
The dashed $t^{-1}$ line is included only as a scaling reference.
The intrinsic kernel reflects the nonuniform relaxation spectrum of
the learned SSM and is not identified with the collective memory
kernel or self-energy dressing associated with the full-block
collective dynamics.
}
\label{fig:intrinsic_ssm_spectrum}
\label{fig:intrinsic_ssm_spectrum}
\end{figure*}

\begin{figure}[t]
\centering
\includegraphics[scale=0.5, trim= 0.4cm 7cm 0cm 0cm]{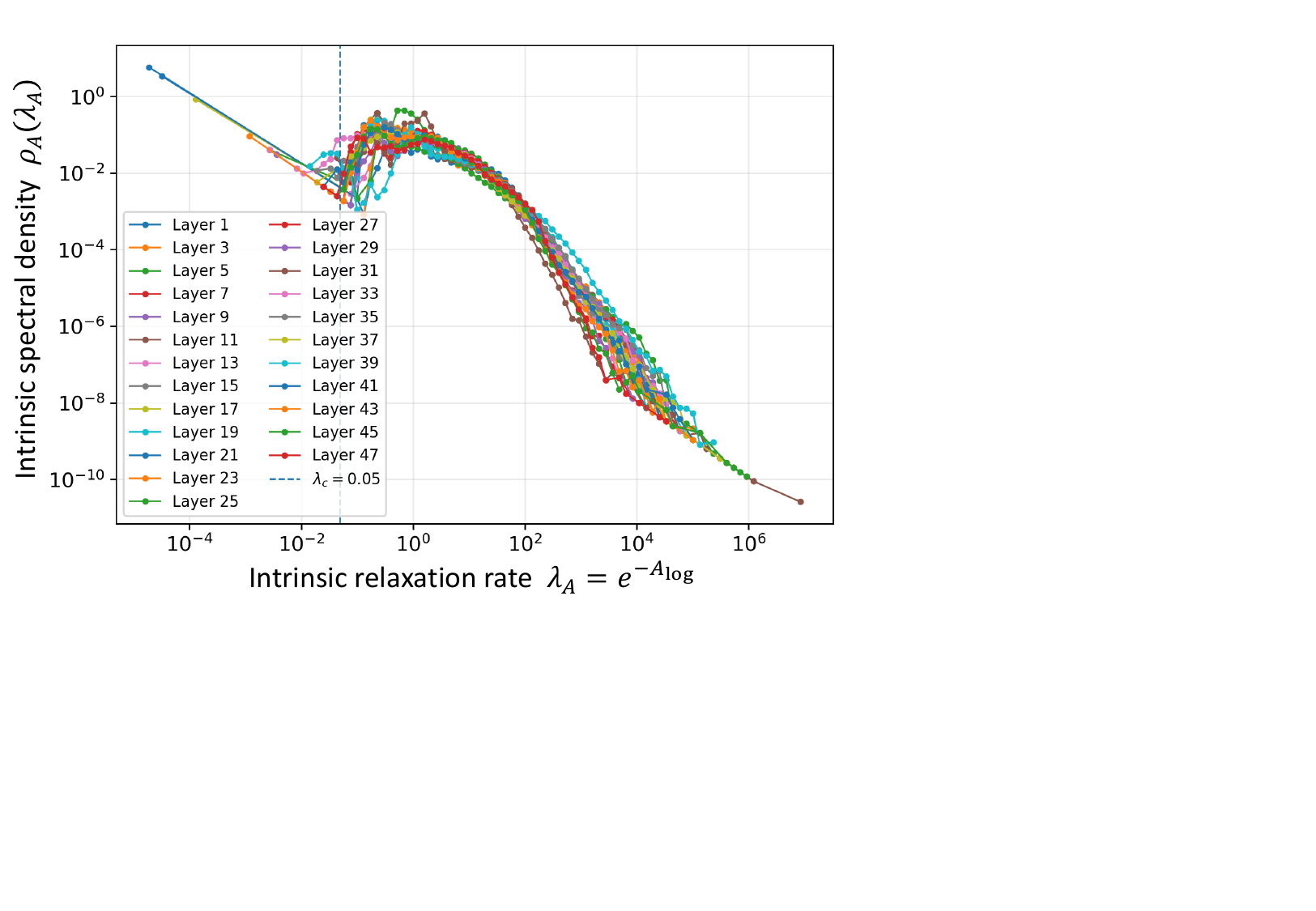}
\caption{
Intrinsic SSM relaxation spectra across all blocks
of Mamba-370M.
The intrinsic relaxation-rate density
$\rho_A(\lambda_A)$, obtained from
$\lambda_A=\exp(A_{\log})$, is shown for all 48 Mamba blocks.
Despite block-to-block variations, the spectra reproduce a common
broad envelope over many orders of magnitude in relaxation rate,
including a distributed slow-relaxation sector extending toward the
deep infrared.
The intermediate and rapidly relaxing sectors exhibit particularly
strong overlap across depth, whereas larger fluctuations appear in
the more sparsely populated infrared region.
The persistence of the overall spectral structure throughout the
network demonstrates that the broad intrinsic relaxation hierarchy is
a distributed property of the learned Mamba state-space architecture
rather than a feature of an isolated block.
}
\label{fig:intrinsic_tdos_layers}
\end{figure}

\subsection{B. Intrinsic relaxation spectrum of the learned SSM}

The full-block spectrum introduced above describes the collective
dynamics after all components of a Mamba block have been combined.
Mamba, however, provides an additional level of dynamical access that
is not generally available in Transformer architectures: the
relaxation spectrum encoded directly in the learned state-space
generator can be examined independently of the complete block
response.

We therefore descend from the full-block collective spectrum to the
intrinsic dynamics of the SSM itself.
In the Mamba parameterization, the stable continuous-time state
generator is represented through learned logarithmic parameters
$A_{\log}$.
The corresponding intrinsic relaxation rates are
\begin{equation}
\lambda_{A,i}
=
\exp(A_{\log,i}),
\label{eq:intrinsic_lambda_A}
\end{equation}
with the negative sign absorbed into the stable continuous-time
generator.
These quantities characterize relaxation scales encoded directly in
the learned SSM before input-dependent discretization, gating, and
the remaining block transformations are applied.

Figure~2(a) shows the learned $A_{\log}$ distribution for a
representative layer.
The distribution spans a broad interval of logarithmic scales,
demonstrating that the learned SSM does not contain a single
characteristic relaxation time.
Through Eq.~(\ref{eq:intrinsic_lambda_A}), this structure corresponds
to a hierarchy of intrinsic relaxation rates extending over many
orders of magnitude.

To characterize this hierarchy, we define the intrinsic SSM spectral
density
\begin{equation}
\rho_A(\lambda_A)
=
\frac{1}{N_A}
\sum_i
\delta(\lambda_A-\lambda_{A,i}),
\label{eq:intrinsic_tdos}
\end{equation}
where $N_A$ is the number of intrinsic SSM modes.
Figure~2(b) shows the resulting spectrum on logarithmic axes.
The resolved intrinsic rates extend approximately from
$\lambda_A\sim10^{-5}$ to beyond $\lambda_A\sim10^{4}$.
Most spectral weight lies at intermediate relaxation rates, while a
lower-density branch extends toward progressively slower modes.
Thus, a broad hierarchy of relaxation scales is already encoded in
the learned state-space generator before selective input conditioning
is applied.

The temporal structure associated with this intrinsic spectrum can be
represented by the spectral relaxation kernel
\begin{equation}
K_A(t)
=
\int_0^\infty
d\lambda_A\,
\rho_A(\lambda_A)e^{-\lambda_A t}
=
\frac{1}{N_A}
\sum_i e^{-\lambda_{A,i}t}.
\label{eq:intrinsic_memory_kernel}
\end{equation}
As shown in Fig.~2(c), the broad intrinsic spectrum produces
relaxation over an extended range of continuous time scales.
The dashed $t^{-1}$ line is included only as a reference; the
intrinsic kernel is not described by a single $1/t$ law over its full
range.

Importantly, $K_A(t)$ is a spectral relaxation kernel of the intrinsic
SSM and should not be identified with the collective memory kernel or
self-energy associated with the full-block dynamics.
It describes the propagation of modes encoded directly in the learned
state-space generator, before selective conditioning and the remaining
nonlinear transformations of the block are included.

A central question is whether this intrinsic slow-mode hierarchy is
specific to the representative layer or distributed throughout the
trained network.
Figure~3 therefore compares the intrinsic spectra across all 48
layers of Mamba-370M.
Across depth, the spectra reproduce a closely related broad envelope
over several decades of relaxation rate.
The intermediate and rapidly relaxing sectors show particularly
strong overlap, whereas the largest layer-to-layer variations occur
in the sparsely populated low-$\lambda_A$ region.

The persistence of this organization throughout the network shows
that the broad intrinsic relaxation hierarchy is not confined to an
isolated layer.
Rather, a distributed hierarchy of intrinsic time scales is encoded
throughout the learned Mamba state-space architecture.
Fluctuations among the slowest modes occur where the spectral
population is sparse and do not alter the common large-scale
organization observed across depth.

The intrinsic spectrum should nevertheless not be equated with the
collective TDOS obtained from the complete block Jacobian.
The intrinsic rates $\lambda_A=\exp(A_{\log})$ are continuous-time
relaxation parameters of the learned SSM generator, whereas the
full-block rates $\lambda_{\rm block}=-\ln|\zeta|$ characterize the
response of the complete hidden-state transformation per block
propagation step.
Their absolute scales, normalization, and spectral forms therefore
need not coincide.

The intrinsic analysis establishes the microscopic starting point of
the relaxation hierarchy,
\begin{equation}
A_{\log}
\longrightarrow
\lambda_A
\longrightarrow
\rho_A(\lambda_A)
\longrightarrow
K_A(t).
\label{eq:intrinsic_spectral_chain}
\end{equation}
The trained Mamba architecture therefore contains a broad,
distributed hierarchy of intrinsic relaxation scales before the
complete block response is constructed.
We next examine how this hierarchy is reorganized by Mamba's
input-dependent selective dynamics.

\begin{figure}[t]
\centering
\includegraphics[scale=0.58, trim= 0.3cm 0.5cm 0cm 0cm]{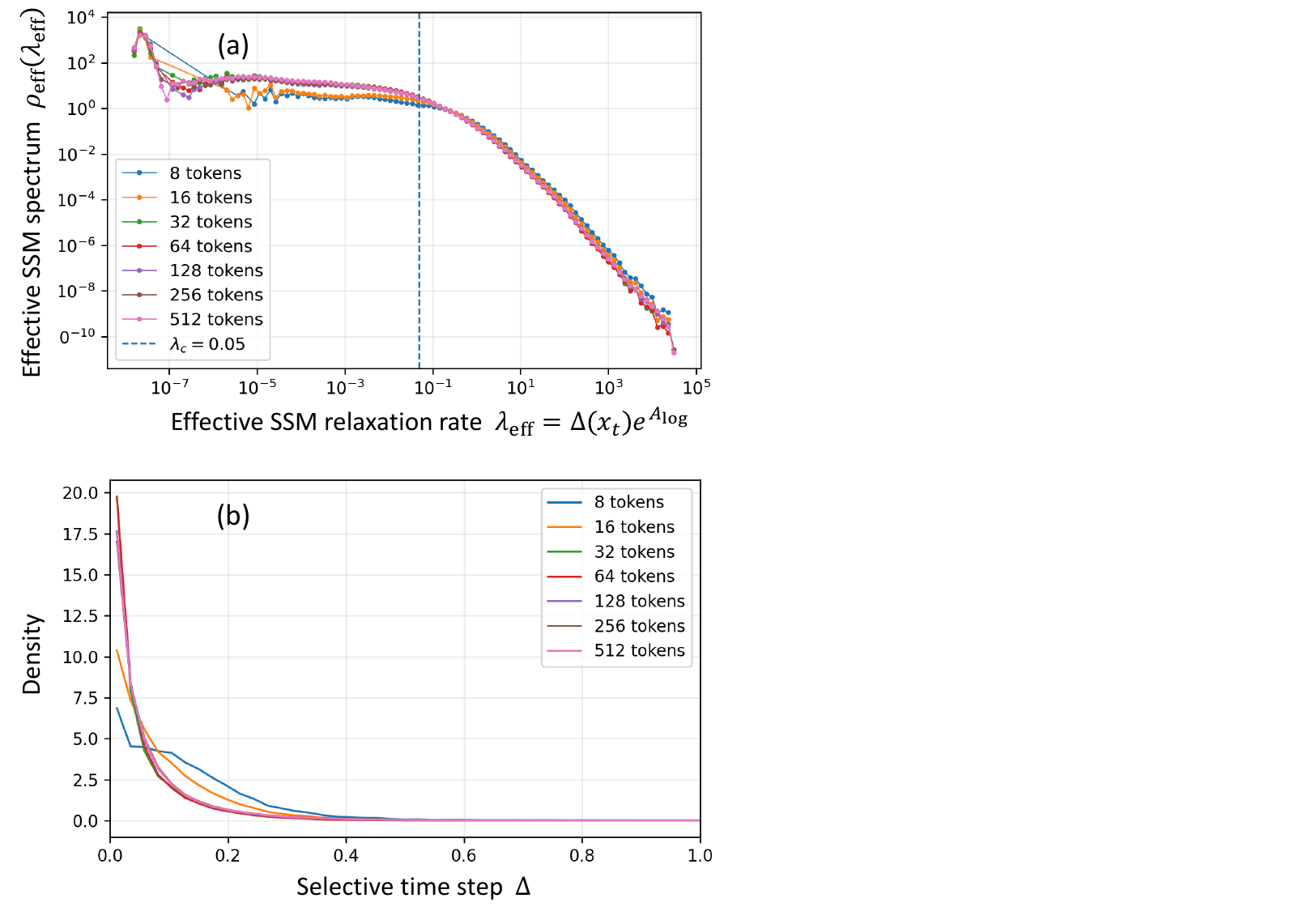}
\caption{
Input-conditioned effective SSM relaxation dynamics in Mamba-370M
at block index $\ell=24$.
(a) Effective relaxation-rate density constructed from the
selectively rescaled rates
$\lambda_{\rm eff}(x_t)=\Delta(x_t)e^{A_{\log}}$
for retained sequence lengths
$L=8,16,32,64,128,256,$ and $512$.
The effective spectrum spans an exceptionally broad hierarchy of
relaxation rates.
Short sequences exhibit substantial fluctuations in the sparsely
sampled infrared sector, whereas increasing sequence length produces
progressively stronger overlap and convergence of the normalized
spectral envelope.
The dashed vertical line at $\lambda_c=0.05$ is shown only as a
common reference for the slow-relaxation sector.
(b) Distribution of the corresponding input-dependent selective
time step $\Delta(x_t)$.
The distributions are strongly weighted toward small $\Delta$ and
become increasingly stable with sequence length.
Because
$\lambda_{\rm eff}=\Delta(x_t)e^{A_{\log}}$,
small selective time steps shift intrinsic SSM modes toward smaller
effective relaxation rates and thereby extend the effective spectrum
toward longer time scales.
Together, the two panels directly show how Mamba's selective
mechanism dynamically rescales the learned intrinsic relaxation
hierarchy during sequence processing.
}
\label{fig:mamba_effective_ssm}
\end{figure}

\subsection{C. Selective temporal rescaling and the effective SSM spectrum}

The intrinsic SSM spectrum analyzed above characterizes the
relaxation structure encoded directly in the learned state-space
generator.
Mamba dynamics, however, does not evolve according to this intrinsic
spectrum alone.
Its local temporal evolution is modulated by the input-dependent
selective time step $\Delta(x_t)$, so that the relaxation scales
activated during inference depend dynamically on the processed input.

For an intrinsic SSM relaxation rate
$\lambda_A=\exp(A_{\log})$, the corresponding effective rate at token
position $t$ is
\begin{equation}
\lambda_{\rm eff}(x_t)
=
\Delta(x_t)\lambda_A
=
\Delta(x_t)\exp(A_{\log}).
\label{eq:effective_ssm_rate}
\end{equation}
The associated relaxation time is
$\tau_{\rm eff}=1/\lambda_{\rm eff}$.
Thus, smaller values of $\Delta(x_t)$ shift a given intrinsic mode
toward slower effective relaxation and correspondingly longer
effective time scales.

To characterize this input-conditioned dynamics over extended
sequences, we construct the effective SSM spectral density from the
collection of token-dependent rates,
\begin{equation}
\rho_{\rm eff}(\lambda)
=
\frac{1}{N_{\rm eff}}
\sum_{t,\alpha}
\delta
\!\left[
\lambda-
\Delta(x_t)\lambda_{A,\alpha}
\right],
\label{eq:effective_ssm_tdos}
\end{equation}
where $\alpha$ labels the intrinsic SSM modes and $t$ labels token
positions.
Unlike the intrinsic spectral density, which is fixed by the learned
SSM parameters, $\rho_{\rm eff}(\lambda)$ incorporates the
input-conditioned temporal modulation generated during inference.

Figure~4(a) shows the effective SSM spectrum of Mamba-370M at layer~24
for sequence lengths ranging from 8 to 512 tokens.
The measured effective rates span an exceptionally broad range,
extending from approximately $10^{-7}$ to beyond $10^{4}$.
These values characterize the resolved input-conditioned SSM spectrum
and should not be interpreted as relaxation rates per full-block
propagation step.

For the shortest sequences, particularly 8 and 16 tokens, the
low-$\lambda_{\rm eff}$ sector is sparsely sampled and exhibits
pronounced finite-sample fluctuations.
As the retained sequence length increases, these fluctuations are
progressively reduced and the broad spectral envelope becomes
increasingly well resolved.
By approximately 32--64 tokens, the large-scale form of the spectrum
is already clearly established, while longer sequences provide
increasingly dense sampling of its slowest effective modes.

At intermediate rates, approximately
$10^{-5}\lesssim\lambda_{\rm eff}\lesssim10^{-2}$, the longer-sequence
spectra show particularly strong overlap and form a broad,
slowly varying region.
At larger rates the spectrum turns downward and subsequently decays
over several decades.
Increasing sequence length therefore primarily improves the
statistical resolution of the effective spectrum rather than
qualitatively changing its overall form.

The convergence of the normalized spectral distribution is important
because increasing sequence length does more than simply increase the
number of sampled effective modes.
The large-scale distribution itself approaches a reproducible form as
longer portions of the sequence are included.
The effective SSM dynamics therefore exhibits a stable statistical
organization over a broad hierarchy of input-conditioned relaxation
scales.

The origin of this temporal rescaling can be examined directly
through the selective time step.
Figure~4(b) shows the measured distributions of $\Delta(x_t)$ for the
same sequence lengths.
The distributions contain substantial weight at small $\Delta$, while
short-sequence fluctuations are progressively reduced as longer
sequences are sampled.

Through Eq.~(\ref{eq:effective_ssm_rate}), small values of $\Delta$
rescale intrinsic modes toward smaller effective relaxation rates.
The selective mechanism therefore provides a direct dynamical route
by which the learned intrinsic relaxation hierarchy can be extended
toward longer effective time scales during inference.
The resulting effective spectrum is determined jointly by the
intrinsic SSM rates and their input-dependent selective rescaling;
selection does not create a new set of intrinsic modes, but
dynamically reorganizes their effective temporal scales.

It is important to distinguish this effective SSM spectrum from the
collective relaxation spectrum of the complete Mamba block.
The dynamical hierarchy is
\begin{equation}
\lambda_A
=
e^{A_{\log}}
\rightarrow
\lambda_{\rm eff}(x_t)
=
\Delta(x_t)e^{A_{\log}}
\rightarrow
\lambda_{\rm block}
=
-\ln|\zeta|,
\label{eq:ssm_to_collective_hierarchy}
\end{equation}
where $\zeta$ denotes an eigenvalue of the full-block Jacobian.
The first quantity characterizes the learned intrinsic SSM generator,
the second its input-conditioned temporal rescaling, and the third
the collective relaxation dynamics of the complete hidden-state
transformation.

The effective SSM spectrum should therefore not be identified with
the collective TDOS.
It represents an intermediate dynamical level connecting the learned
state-space generator to the collective response of the complete
Mamba block.
Mamba thus allows the microscopic relaxation hierarchy and its
selective temporal rescaling to be resolved explicitly before the
resulting dynamics is reorganized at the full-block level.
We next examine this collective spectrum and its infrared
organization.

\begin{figure}[t]
\centering
\includegraphics[scale=0.5, trim= 0.3cm 7cm 0cm 0cm]{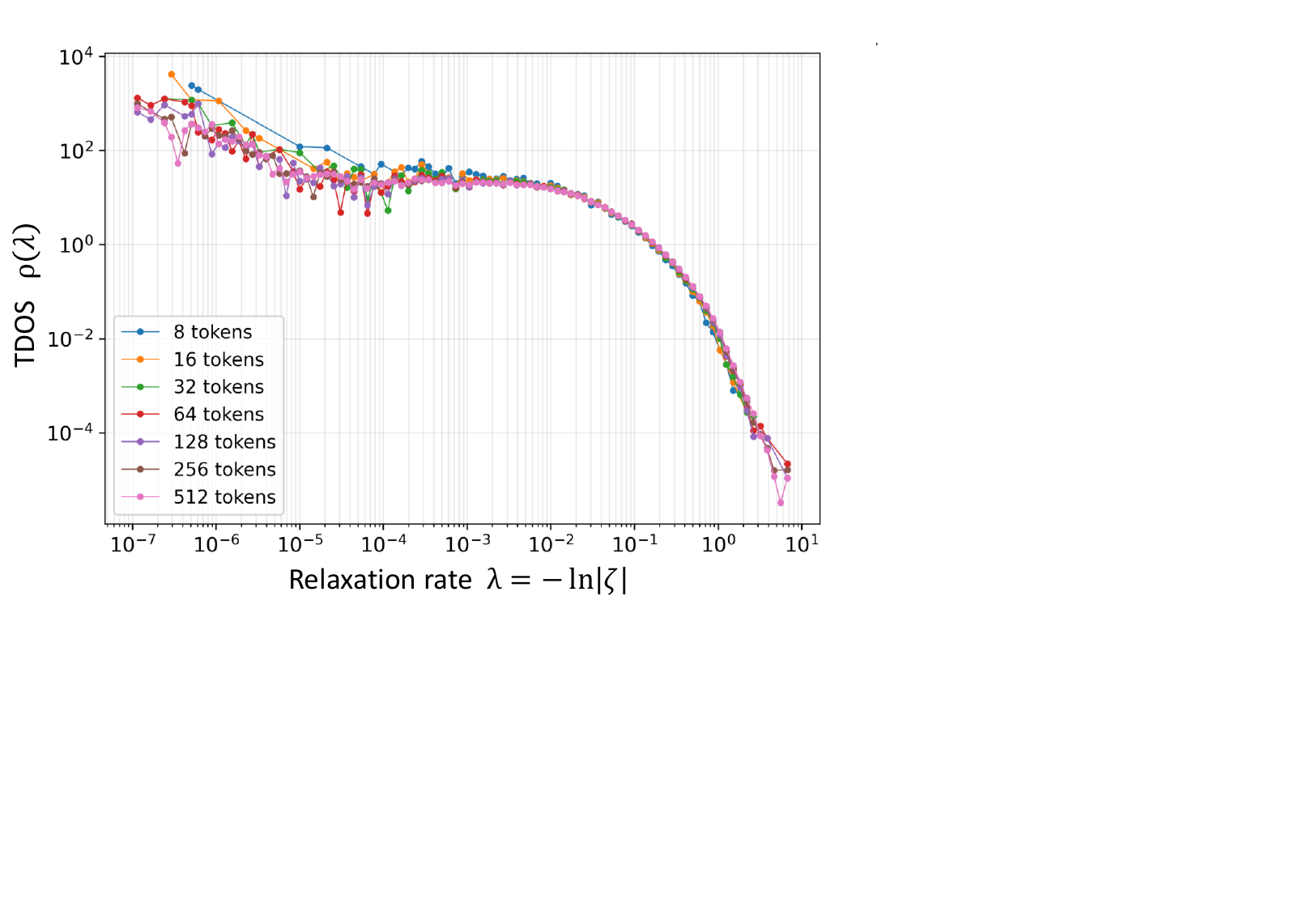}
\caption{
Sequence-length dependence of the full-block collective relaxation
spectrum in Mamba-370M.
The time-scale density of states is obtained from the eigenvalues
$\zeta_\alpha$ of the complete block Jacobian through
$\lambda_\alpha=-\ln|\zeta_\alpha|$ and is shown for sequence lengths
$L=8$--$512$ at layer 24.
The measured collective spectrum spans approximately eight decades in
relaxation rate, from $\lambda\sim10^{-7}$ to $\lambda\sim10^{1}$.
Short sequences exhibit visible finite-size fluctuations in the
deep-infrared sector, whereas increasing sequence length progressively
resolves slower modes and produces a smoother common spectral envelope.
For $L=512$, the collective spectral profile remains comparatively
smooth down to approximately $\lambda\sim5\times10^{-6}$, while
individual modes extend further into the sparsely populated
extreme-infrared regime near $\lambda\sim10^{-7}$.
The simultaneous extension and convergence of the low-$\lambda$ sector
indicate that the infrared continuum becomes progressively better
resolved as a larger dynamical subspace is sampled.
}
\label{fig:mamba_fullblock_sequence_tdos}
\end{figure}

\begin{figure}[t]
\centering
\includegraphics[scale=0.5, trim= 0.3cm 7cm 0cm 0cm]{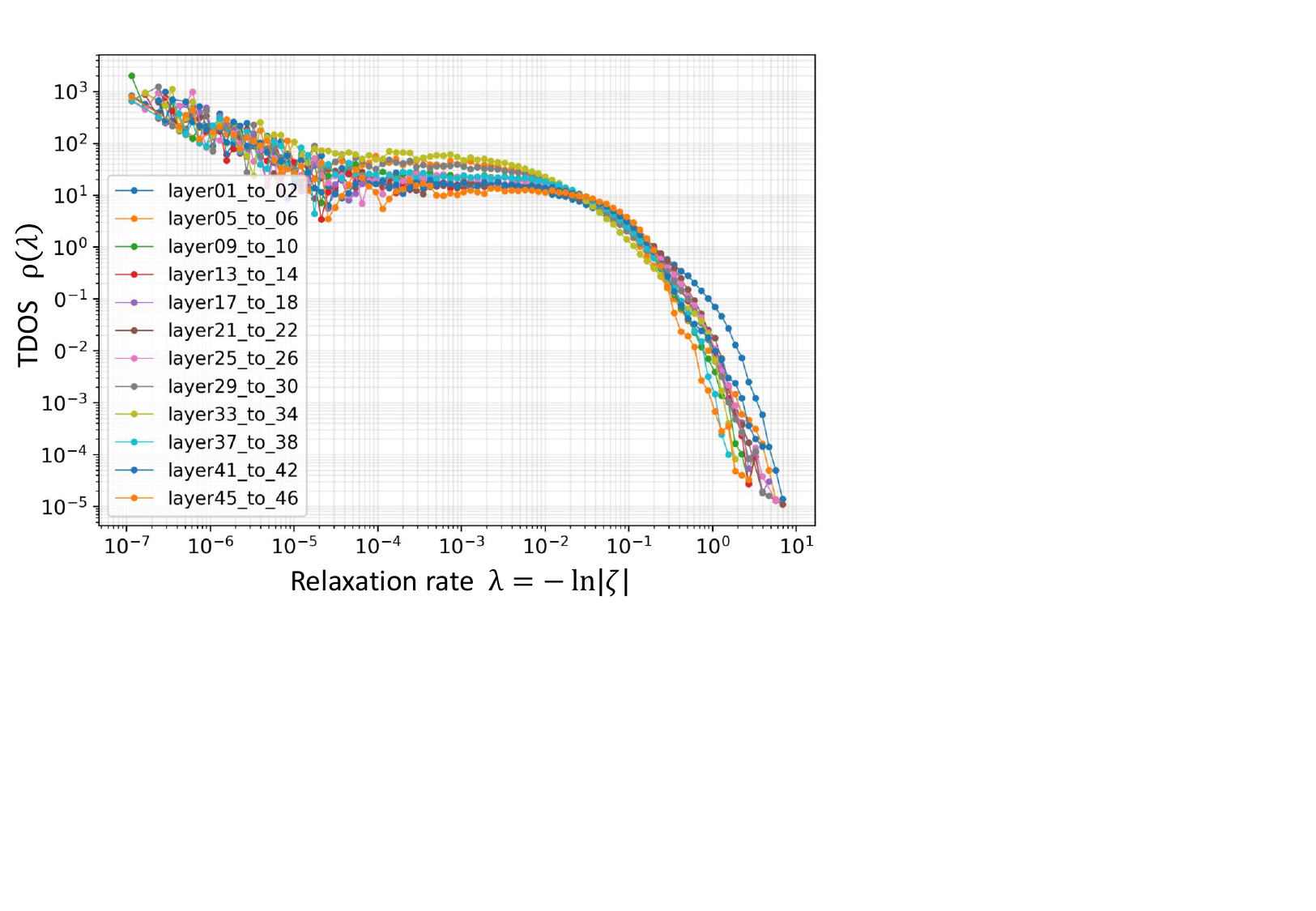}
\caption{
Depth dependence of the full-block collective relaxation spectrum in
Mamba-370M.
The time-scale density of states is measured from full-block
Jacobians at a fixed sequence length of $L=128$ for layer pairs
distributed throughout the 48-layer network, from layers $1\!\to\!2$
to $45\!\to\!46$.
Despite quantitative variations in local spectral weight and in the
high-relaxation-rate rolloff, the measured spectra reproduce a closely
related organization across depth: a broad slow-relaxation sector,
an extended weakly varying infrared region, and a rapid decay toward
the ultraviolet.
The collective spectra extend over multiple decades into the
deep-infrared regime, while the overall spectral envelope remains
similar across widely separated layers.
The persistence of this low-$\lambda$ organization across widely
separated depths indicates that the collective infrared structure is
not restricted to a particular layer but is distributed throughout
the trained Mamba model.
}
\label{fig:mamba_fullblock_depth_tdos}
\end{figure}

\subsection{D. Emergence of the full-block collective relaxation spectrum}

The intrinsic and effective SSM spectra characterize the internal
relaxation structure of the state-space component of Mamba.
The complete block, however, additionally contains input-dependent
projections, gating, nonlinear transformations, channel mixing, and
residual propagation.
Its collective relaxation dynamics must therefore be determined from
the full hidden-state transformation rather than from the SSM
generator alone.

Using the full-block Jacobian construction introduced in Sec.~III.A,
we obtain the collective relaxation rates
$\lambda_\alpha=-\ln|\zeta_\alpha|$ and the corresponding
full-block TDOS $\rho_{\rm block}(\lambda)$.
This spectrum constitutes the final collective level of the
intrinsic--selective--collective hierarchy established above and
provides the direct counterpart of the collective TDOS measured from
Transformer block Jacobians.

Figure~5 shows the full-block TDOS at layer~24 for retained
sequence lengths from 8 to 512 tokens.
A broad collective relaxation spectrum is observed over many decades,
with a substantial population of modes extending toward small
relaxation rates.
The intermediate and rapidly relaxing parts of the spectrum approach
a common envelope already at relatively short sequence lengths,
whereas the strongest sequence-length dependence occurs in the
infrared sector.

For short sequences, the lowest-relaxation-rate region is sparsely
sampled and exhibits substantial finite-sampling fluctuations.
Increasing the retained sequence length progressively fills this
region, suppresses the fluctuations, and extends the smoothly resolved
collective spectrum toward smaller relaxation rates.
For the eight-token measurement, substantial fluctuations appear
below approximately $\lambda\sim10^{-4}$--$10^{-3}$, whereas for
512 tokens the spectral profile remains comparatively smooth down to
approximately $\lambda\sim5\times10^{-6}$.
The reliably resolved infrared continuum therefore expands
substantially as the measured sequence subspace is enlarged.

This behavior should be distinguished from the occurrence of isolated
eigenvalues at still smaller relaxation rates.
Such modes extend to values of order $\lambda\sim10^{-7}$, but the
extreme infrared remains sparsely populated and correspondingly more
sensitive to finite-sampling fluctuations.
The relevant convergence is instead the progressive transformation of
this fluctuating region into a densely sampled and reproducible
spectral continuum.
Increasing sequence length therefore improves the infrared resolution
while preserving the common large-scale spectral envelope.

Across the resolved slow-relaxation sector, the TDOS varies only
weakly over a broad range and develops enhanced spectral weight toward
smaller $\lambda$, before crossing over to rapid suppression at larger
relaxation rates.
The numerical crossover scales are approximate features of the
measured spectra rather than imposed spectral boundaries.
The histogram itself is used here primarily to establish the
morphology and sequence-length convergence of the collective
infrared sector; its scaling exponent is determined independently
below from cumulative mode statistics.

The robustness of this organization can be tested independently
across network depth.
Figure~6 compares full-block TDOS measurements from representative
layers distributed throughout Mamba-370M at a fixed sequence length
of 128 tokens.
Although the detailed spectral weight and local structure vary with
depth, the same large-scale morphology is repeatedly recovered:
a broad slow-relaxation continuum, a weakly varying infrared sector,
and rapid suppression toward the ultraviolet.

The full-block measurements therefore reveal two complementary forms
of robustness.
Increasing sequence length progressively improves the statistical
resolution of the same infrared continuum and leads toward a stable
large-scale spectral form, while measurements across network depth
recover the same qualitative collective organization from distinct
block transformations.
Together, these observations indicate that the slow-mode spectrum is
a distributed property of the trained Mamba hidden-state dynamics
rather than a feature of a particular token subspace or isolated
layer.

The result also completes the microscopic-to-macroscopic hierarchy
resolved by the Mamba architecture.
The learned SSM provides a broad intrinsic relaxation substrate,
input-dependent selection dynamically rescales its temporal structure,
and the complete nonlinear block reorganizes these dynamics into the
collective TDOS measured from the hidden-state Jacobian.
The remaining question is whether the resolved collective infrared
sector is described by a reproducible scaling law and whether this
scaling itself converges as the measured sequence subspace is
enlarged.
We address this next using cumulative mode statistics, avoiding direct
dependence on histogram binning.

\begin{figure*}[t]
\centering
\includegraphics[width=0.9\textwidth, trim=0cm 4.5cm 0cm 0cm]{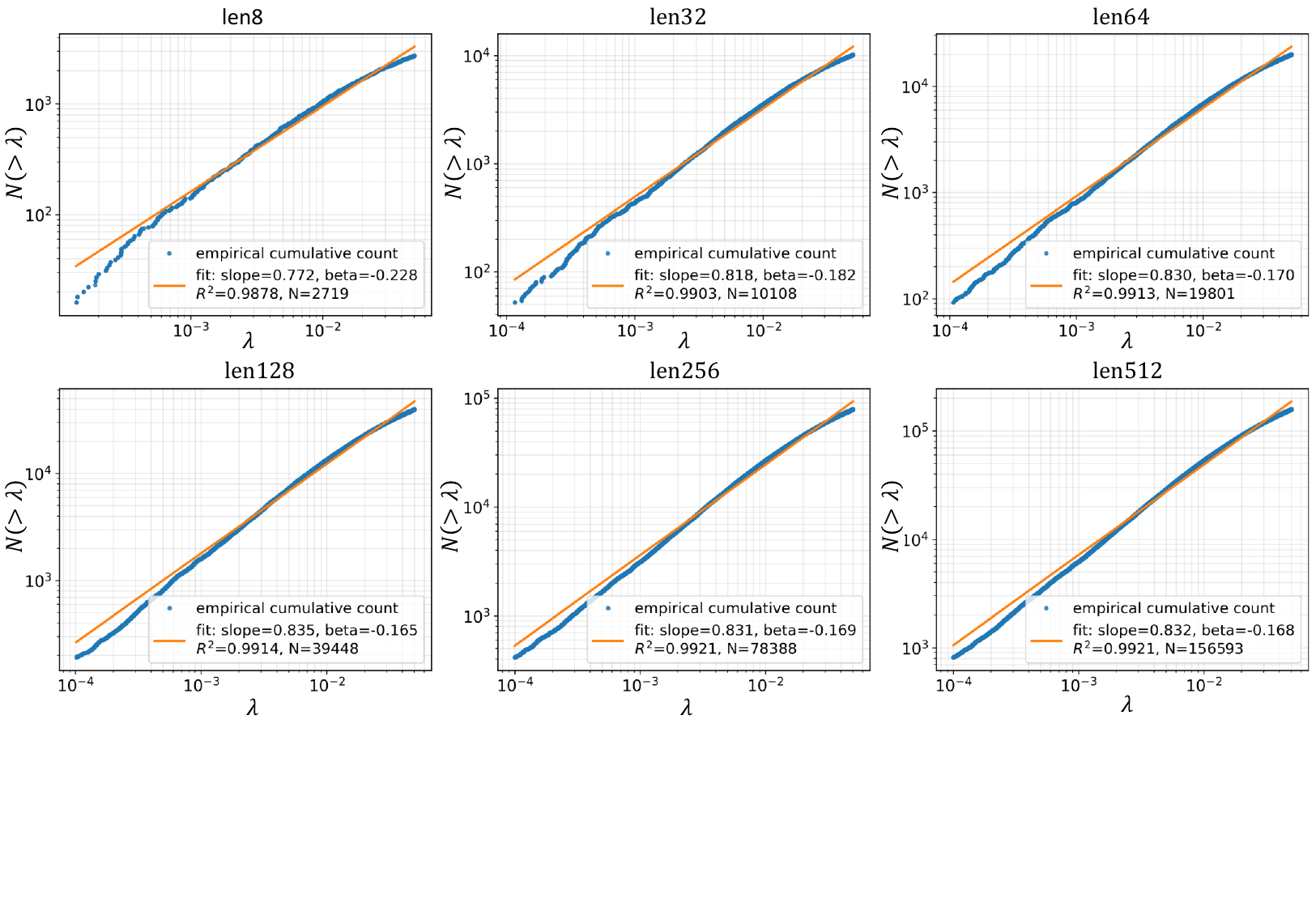}
\caption{
Cumulative infrared scaling of the full-block relaxation spectrum in
Mamba-370M.
The empirical cumulative number of stable relaxation modes,
$N(<\lambda)$, is shown at layer~24 for sequence lengths
$L=8,16,32,64,128,256,$ and $512$.
For each sequence length, the cumulative spectrum is fitted over the
same fixed infrared interval,
$10^{-4}\leq\lambda\leq5\times10^{-2}$, according to
$N(<\lambda)\propto\lambda^s$, with the corresponding TDOS exponent
given by $\beta=s-1$.
The fitted exponent evolves from $\beta=-0.228$ at $L=8$ and
stabilizes near $\beta\simeq-0.17$ for the longer sequences, while
the coefficients of determination remain approximately
$R^2\simeq0.99$ throughout.
Over the same measurements, the number of modes within the fixed
fitting interval increases from $N_{\rm fit}=2,719$ at $L=8$ to
$156,593$ at $L=512$.
Thus, the infrared statistics improve by nearly a factor of 58 while
the cumulative scaling exponent remains essentially unchanged at the
largest sequence lengths.
The more sparsely sampled deep-infrared region below the common
fitting interval is excluded from the exponent determination.
}
\label{fig:mamba_cumulative_ir_scaling}
\end{figure*}

\subsection{E. Infrared scaling of the full-block collective spectrum}

The full-block measurements above reveal a broad slow-relaxation
continuum whose resolved infrared extent increases systematically with
sequence length.
We now test whether this collective spectrum exhibits a reproducible
infrared scaling law.

Direct power-law fitting of histogram-based TDOS estimates is
sensitive to logarithmic binning, particularly in the sparsely sampled
deep-infrared region.
We therefore determine the infrared exponent from the empirical
cumulative mode distribution.
For
\begin{equation}
\rho(\lambda)
\sim
C\lambda^\beta,
\label{eq:mamba_tdos_powerlaw}
\end{equation}
the cumulative number of modes obeys
\begin{equation}
N(<\lambda)
\propto
\lambda^{1+\beta},
\end{equation}
for $\beta>-1$.
Thus, writing $N(<\lambda)\propto\lambda^s$ gives
\begin{equation}
\beta=s-1.
\label{eq:mamba_beta_from_cdf}
\end{equation}

To compare different sequence lengths on the same basis, all
cumulative spectra are fitted over the fixed interval
\begin{equation}
10^{-4}
\leq
\lambda
\leq
5\times10^{-2}.
\label{eq:mamba_fit_window}
\end{equation}
This conservative interval lies within the well-resolved
slow-relaxation sector common to all sequence lengths and is held
fixed independently of $L$.
The more sparsely sampled deep-infrared region below this interval is
therefore excluded from the exponent extraction, even though the
longer-sequence measurements resolve the collective spectrum to
substantially smaller relaxation rates.

Figure~7 shows the cumulative distributions and corresponding fits
for sequence lengths from 8 to 512 tokens.
The spectra are approximately linear on logarithmic coordinates over
the common fitting window, with coefficients of determination near
$R^2\simeq0.99$ throughout.
The quantitative results are summarized in Table~I.

\begin{table}[t]
\centering
\caption{
Infrared scaling of the Mamba-370M full-block Jacobian TDOS at
layer~24.
All cumulative spectra are fitted over the fixed interval
$10^{-4}\leq\lambda\leq5\times10^{-2}$ according to
$N(<\lambda)\propto\lambda^s$, with $\beta=s-1$.
$N_{\rm fit}$ denotes the number of modes within the fitting interval.
}
\label{tab:mamba_ir_exponents}

\begin{tabular*}{\columnwidth}
{@{\extracolsep{\fill}}ccccc@{}}
\hline\hline
Tokens & $s$ & $\beta$ & $R^2$ & $N_{\rm fit}$ \\
\hline
8   & 0.772 & -0.228 & 0.988 & 2,719   \\
16  & 0.802 & -0.198 & 0.989 & 5,194   \\
32  & 0.818 & -0.182 & 0.990 & 10,108  \\
64  & 0.830 & -0.170 & 0.991 & 19,801  \\
128 & 0.835 & -0.165 & 0.991 & 39,448  \\
256 & 0.831 & -0.169 & 0.992 & 78,388  \\
512 & 0.832 & -0.168 & 0.992 & 156,593 \\
\hline\hline
\end{tabular*}
\end{table}

Two features are particularly important.
First, the fitted exponent is negative for every sequence length,
ranging from $\beta=-0.228$ to approximately $-0.165$.
The resolved collective spectrum therefore lies consistently on the
infrared-enhanced side of the flat-TDOS limit, while the small
magnitude $|\beta|\ll1$ indicates that the spectrum remains close to
flat.

Second, the exponent becomes increasingly stable as the resolved
sequence subspace is enlarged.
As summarized in Fig.~8, the values for $L=64$, 128, 256, and 512
are approximately $-0.170$, $-0.165$, $-0.169$, and $-0.168$,
respectively.
Thus, despite an eightfold increase in sequence length from 64 to
512 tokens, the fitted exponent remains confined to a narrow interval
of width less than $0.01$.
Within the sequence lengths examined here, the resolved infrared
exponent therefore stabilizes near
\begin{equation}
\beta_{\rm IR}
\simeq
-0.17.
\label{eq:mamba_beta_stable}
\end{equation}
The persistence of this value through the 512-token measurement also
shows that the observed evolution is not a simple monotonic drift
toward the exactly flat value $\beta=0$.

This stabilization occurs while the number of modes contained in the
same fixed fitting interval increases from $2,719$ at $L=8$ to
$156,593$ at $L=512$, an increase by nearly a factor of 58.
Increasing sequence length therefore greatly improves the infrared
statistics while leaving the fitted scaling exponent essentially
unchanged over the largest measured sequence subspaces.
Together with the progressive extension and smoothing of the directly
measured TDOS in Fig.~5, this behavior provides strong evidence that
the weakly negative exponent is not simply produced by sparse
short-sequence sampling.

Because the exponent is obtained from the slope of the cumulative
distribution rather than the absolute number of modes, increasing the
Jacobian dimension changes the available statistics without trivially
fixing the measured value of $\beta$.
The persistence of nearly the same cumulative slope while the sampled
mode population increases by almost two orders of magnitude therefore
provides an independent test of the relative organization of the
collective relaxation spectrum.

The dynamical implication follows from the previously established
relation $K(t)\sim t^{-(1+\beta)}$.
The stabilized value $\beta_{\rm IR}\simeq-0.17$ therefore corresponds
to
\begin{equation}
K(t)\sim t^{-0.83}
\end{equation}
over the dynamical regime associated with the resolved infrared
scaling window.
The full-block dynamics consequently lies close to the marginal
$1/t$ memory form, with a weak enhancement of long-time persistence.

Taken together, the cumulative fits, the consistently negative
infrared exponent, its stabilization with increasing sequence length,
and the simultaneous growth of infrared statistics identify a
reproducible scaling regime in the Mamba full-block dynamics.
The collective spectrum is therefore most naturally characterized,
over the resolved infrared window, as a nearly flat but weakly
infrared-enhanced TDOS with $\beta_{\rm IR}\simeq-0.17$.
This exponent characterizes the measured scaling regime and is not
intended as an extrapolation to the asymptotic
$\lambda\rightarrow0$ limit.

\begin{figure}[t]
\centering
\includegraphics[scale=0.48, trim= 0.3cm 7cm 0cm 0cm]{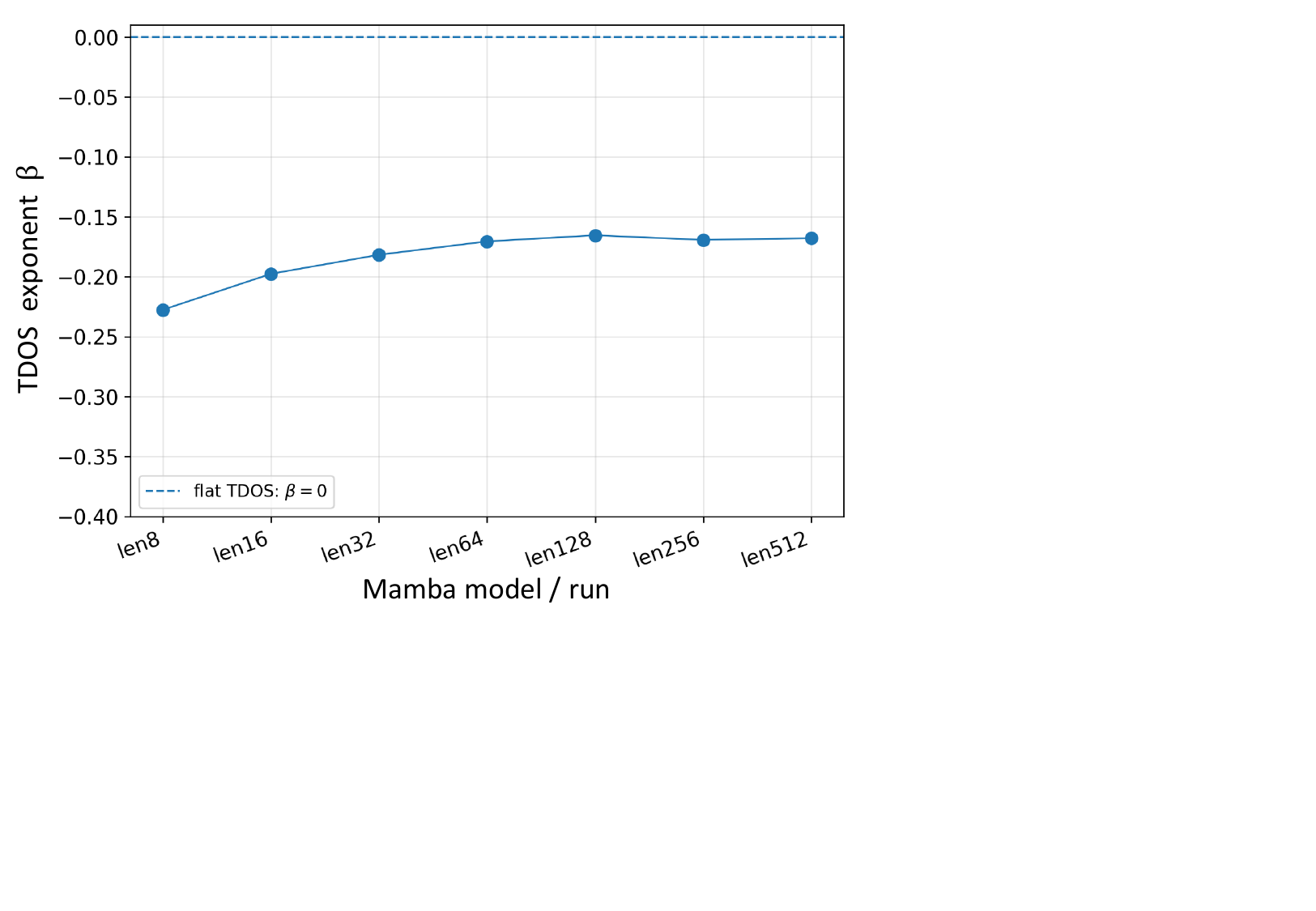}
\caption{
Sequence-length dependence of the infrared TDOS exponent in the
Mamba-370M full-block dynamics.
The exponent $\beta$ is obtained from the cumulative scaling relation
$N(<\lambda)\propto\lambda^{1+\beta}$ over the common fitting interval
$10^{-4}\leq\lambda\leq5\times10^{-2}$.
As the retained sequence length increases from 8 to 512 tokens,
$\beta$ evolves from $-0.228$ toward a stable value near $-0.17$.
For the four longest sequence lengths,
$L=64,128,256,$ and $512$, the measured values are approximately
$\beta=-0.170,-0.165,-0.169,$ and $-0.168$, respectively.
Thus, despite an eightfold increase in sequence length from 64 to
512 tokens, the fitted exponent remains confined to a narrow interval
below the exactly flat-TDOS limit.
The dashed horizontal line denotes this flat-spectrum limit,
$\beta=0$.
The stabilization of the measured exponent near
$\beta_{\rm IR}\simeq-0.17$, without a systematic drift toward zero,
indicates a reproducible, nearly flat but weakly infrared-enhanced
scaling regime over the resolved infrared window.
}
\label{fig:mamba_beta_sequence_length}
\end{figure}

\begin{figure*}[t]
\centering
\includegraphics[width=1.0\textwidth, trim=0cm 1cm 0cm 0cm]{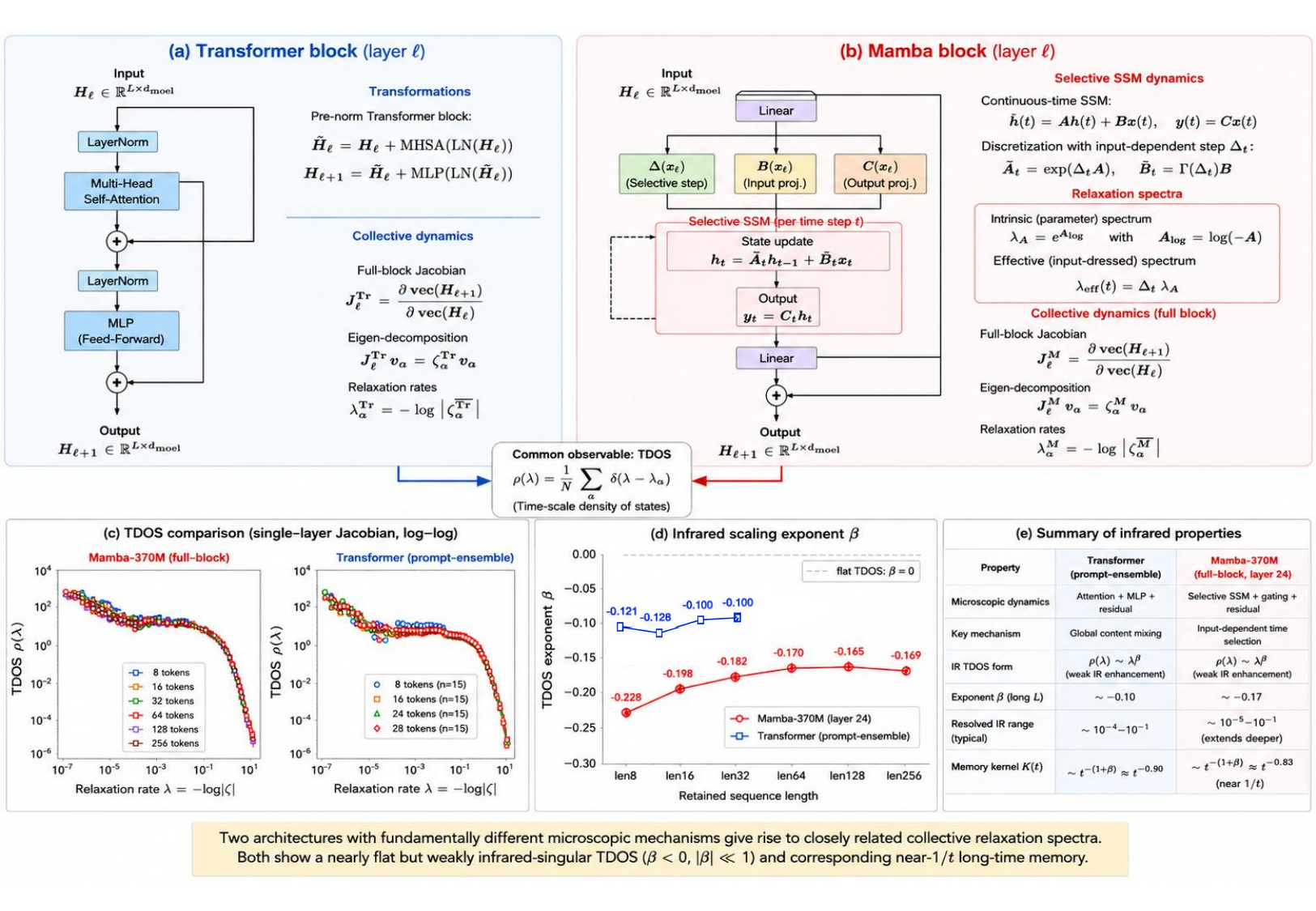}
\caption{
Comparison of the microscopic architectures and collective infrared
dynamics of Transformer and Mamba.
The two architectures generate hidden-state dynamics through
fundamentally different computational mechanisms:
Transformer blocks combine attention, nonlinear feed-forward
transformations, and residual propagation, whereas Mamba blocks employ
selective state-space evolution with input-dependent temporal
modulation.
Despite this difference, both architectures can be analyzed using the
same full-block collective construction.
For each block, the Jacobian of the complete hidden-state
transformation is diagonalized and the collective relaxation rates are
defined as
$\lambda_\alpha=-\ln|\zeta_\alpha|$, allowing the corresponding
time-scale density of states $\rho(\lambda)$ to be compared on the
same dynamical convention.
The measured full-block spectra exhibit a closely related infrared
organization: both contain a broad slow-mode continuum with a nearly
flat but weakly enhanced TDOS toward small relaxation rates.
Power-law analysis,
$\rho(\lambda)\sim\lambda^\beta$, yields weakly negative infrared
exponents in both architectures.
The Transformer measurements approach values of order
$\beta_{\rm Tr}\sim-0.1$, whereas the long-sequence Mamba measurements
stabilize near $\beta_{\rm M}\sim-0.17$.
The numerical exponents are not identical; rather, the common feature
is the regime $\beta<0$ with $|\beta|\ll1$.
Thus, substantially different microscopic architectures produce
closely related near-marginal collective infrared dynamics when
examined through the complete hidden-state transformation.
}
\label{fig:transformer_mamba_collective_comparison}
\end{figure*}

\begin{figure*}[t]
\centering
\includegraphics[width=0.95\textwidth, trim=0cm 0.3cm 1cm 0cm]{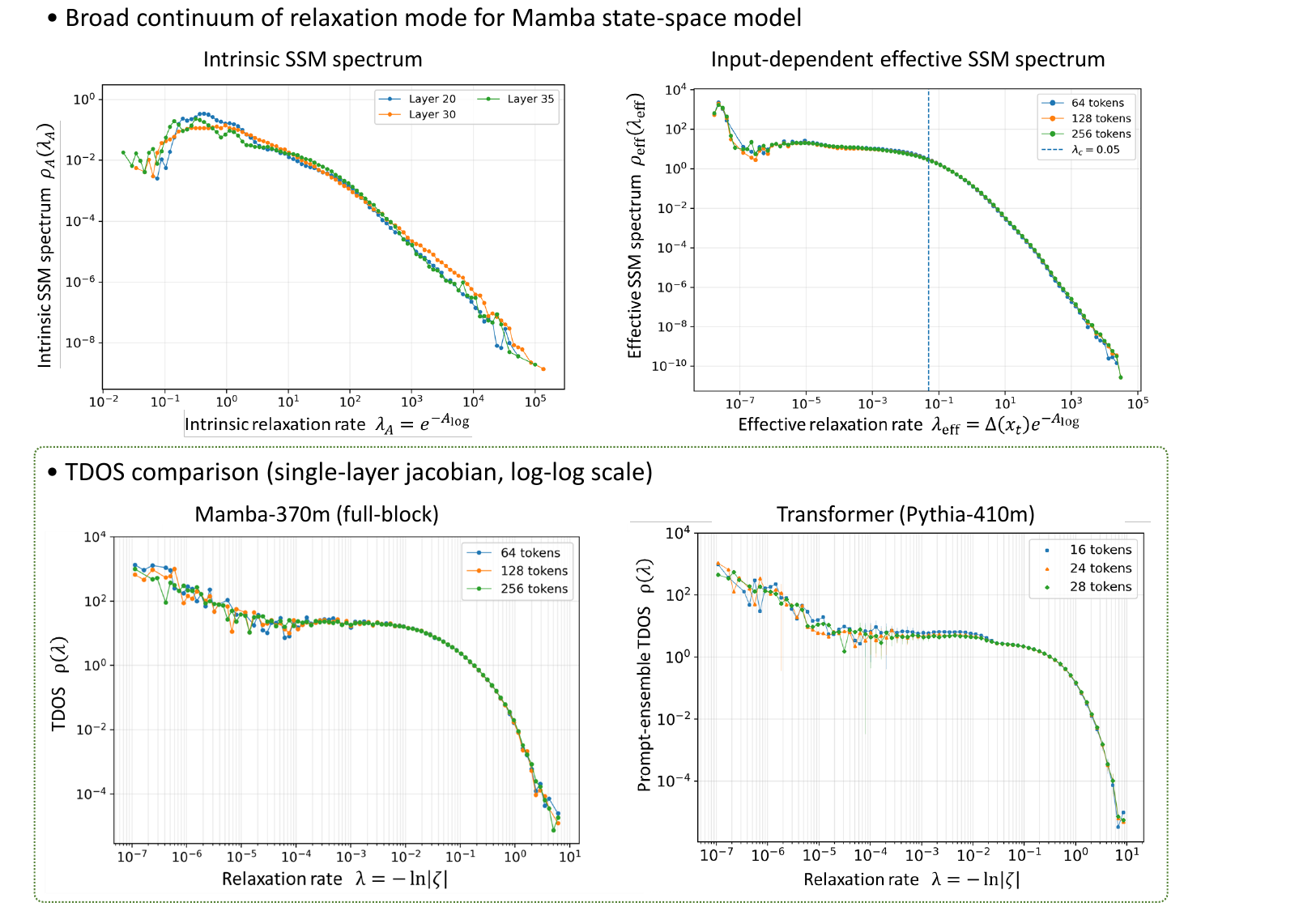}
\caption{
Hierarchy from microscopic state-space relaxation to collective
infrared organization in Mamba, together with the corresponding
collective Transformer spectrum.
The learned Mamba SSM first defines an intrinsic distribution of
relaxation rates,
$\lambda_A=\exp(A_{\log})$.
Input-dependent selective dynamics then temporally rescales these
intrinsic modes according to
$\lambda_{\rm eff}(t)=\Delta_t\lambda_A$, producing an effective SSM
spectrum conditioned on the processed sequence.
After input and output projections, gating, nonlinear transformations,
and residual propagation are included, the complete Mamba block is
characterized independently by the eigenvalues of its full-block
Jacobian and the collective rates
$\lambda=-\ln|\zeta|$.
These three spectra represent distinct dynamical levels and should not
be interpreted as identical spectral objects.
Rather, the comparison shows that a broad slow-mode hierarchy present
in the internal state-space dynamics remains manifest, after
substantial spectral reorganization, in the collective response of the
complete Mamba block.
The Transformer full-block spectrum, obtained independently from the
same Jacobian construction, exhibits a similarly broad and nearly
flat infrared collective sector despite the absence of an explicit
state-space relaxation hierarchy.
The figure therefore illustrates the separation between microscopic
dynamical implementation and the closely related infrared
organization observed at the collective level.
}
\label{fig:microscopic_to_collective_ir}
\end{figure*}

\section{IV. Architecture-Independent Infrared Collective Organization}

The preceding analysis establishes that Mamba develops a broad
continuum of collective relaxation modes extending into the infrared.
This organization can be traced from the intrinsic SSM relaxation
spectrum, through input-dependent selective temporal rescaling, to the
collective spectrum of the complete nonlinear block.
We now ask whether the resulting infrared organization is specific to
the state-space architecture of Mamba or reflects a more general
property of learned sequence dynamics.

This question can be addressed by direct comparison with Transformers.
The comparison is particularly informative because the microscopic
dynamics of the two architectures are fundamentally different.
A Transformer block organizes its hidden representation through
attention, nonlinear feed-forward transformations, and residual
propagation, schematically,
\begin{align}
\widetilde H_\ell
&=
H_\ell+
\mathrm{Attn}
\!\left[
\mathrm{Norm}(H_\ell)
\right],
\\
H_{\ell+1}
&=
\widetilde H_\ell+
\mathrm{MLP}
\!\left[
\mathrm{Norm}(\widetilde H_\ell)
\right].
\end{align}

Mamba instead contains an explicit selective state-space dynamics,
\begin{equation}
h_t
=
\overline A_t h_{t-1}
+
\overline B_t x_t,
\qquad
\overline A_t
=
\exp(\Delta_t A),
\end{equation}
in which the input-dependent selective time step $\Delta_t$ locally
rescales the internal state evolution.

Despite these distinct microscopic mechanisms, both architectures can
be analyzed through the same collective observable at the level of the
complete hidden-state transformation.
For either architecture, we construct the full-block Jacobian
\[
J_\ell
=
\frac{
\partial\,\mathrm{vec}(H_{\ell+1})
}{
\partial\,\mathrm{vec}(H_\ell)
},
\]
with eigenvalues
$J_\ell v_\alpha=\zeta_\alpha v_\alpha$.
Taking propagation through one complete block as one discrete
dynamical step defines the common relaxation coordinate
\[
\lambda_\alpha
=
-\ln|\zeta_\alpha|.
\]
The resulting distribution of collective relaxation modes defines the
full-block time-scale density of states $\rho(\lambda)$.

Figure~9 summarizes this architecture-level comparison.
The use of a common full-block observable does not imply equivalence
of the underlying computations.
In Mamba, the learned SSM generator provides an intrinsic relaxation
hierarchy, which is selectively rescaled by the input-dependent time
step and subsequently reorganized by gating, projections,
nonlinearities, and residual propagation.
Transformer reaches the same level of collective description through
a fundamentally different microscopic route.
The full-block Jacobian therefore provides a common dynamical
coordinate in which the collective organization of the two
architectures can be compared without identifying their internal
mechanisms.

Figure~10 presents the corresponding empirical comparison.
Although the detailed spectral densities and ultraviolet structures
remain architecture dependent, Transformer and Mamba exhibit closely
related organization in the low-relaxation-rate sector.
Both contain a broad continuum of collective modes extending toward
small $\lambda$, together with an extended infrared region whose TDOS
is nearly flat but weakly enhanced toward slower relaxation rates.

We quantify this organization by parameterizing the resolved infrared
spectrum as
\begin{equation}
\rho(\lambda)
\sim
C\lambda^\beta,
\label{eq:architecture_beta}
\end{equation}
where $\beta=0$ represents a flat TDOS and $\beta<0$ corresponds to an
enhancement of spectral weight toward the infrared.

The measured exponents are not numerically identical.
For the Transformer comparison, we use the Pythia-410M model analyzed
in our previous study~\cite{1,18}, in which the collective relaxation
spectrum was measured from the Jacobian of the complete hidden-state
transformation using the same full-block spectral construction.
The resolved Pythia-410M values are approximately
$-0.121$, $-0.128$, $-0.10$, and $-0.10$ for retained sequence
lengths $L=8$, $16$, $24$, and $28$, respectively, giving a
representative infrared exponent of order
\begin{equation}
\beta_{\rm Tr}\sim-0.1.
\end{equation}

For Mamba-370M, the exponent evolves from $\beta=-0.228$ at $L=8$
toward values clustered near $-0.17$ for the longer measurements,
with $\beta=-0.170$, $-0.165$, and $-0.169$ at $L=64$, $128$, and
$256$, respectively.
The long-sequence Mamba measurements are therefore characterized by
\begin{equation}
\beta_{\rm M}\sim-0.17.
\end{equation}

The numerical difference between these values is not unexpected.
The architectures, trained models, microscopic dynamics, and resolved
sequence-length ranges are different.
More important is that both spectra remain close to the flat-TDOS
limit on its weakly infrared-enhanced side.
The sequence-length dependence further supports this interpretation:
increasing the resolved dynamical subspace improves the infrared
statistics without driving either architecture toward an
infrared-depleted spectrum.
In Mamba, the exponent becomes nearly stationary for the longest
sequences while the directly measured TDOS becomes progressively
smoother.
The Transformer measurements likewise remain within the weakly
negative near-flat regime over the sequence lengths examined.

The significance of this similarity lies precisely in the difference
between the microscopic mechanisms that generate the two collective
spectra.
The comparison can therefore be summarized as
\begin{equation}
\begin{aligned}
\text{Transformer:}\qquad
&
\text{attention/MLP dynamics}
\longrightarrow
\rho_{\rm Tr}(\lambda),
\\
\text{Mamba:}\qquad
&
\text{selective SSM dynamics}
\longrightarrow
\rho_{\rm M}(\lambda).
\end{aligned}
\label{eq:architecture_routes}
\end{equation}
The microscopic paths are different, yet both terminate in broad,
nearly flat collective relaxation spectra with weak infrared
enhancement.
Moreover, as established in Sec.~III, the Mamba full-block TDOS is
not a direct copy of either its intrinsic or effective SSM spectrum;
substantial spectral reorganization occurs before the collective
block response is formed.
The similarity observed in Fig.~10 therefore emerges at the
collective rather than microscopic level.

Quantitatively, the representative values
$\beta_{\rm Tr}\sim-0.1$ and $\beta_{\rm M}\sim-0.17$ are not
interpreted as a common universal exponent.
Rather, both architectures occupy the same qualitative infrared
regime,
\begin{equation}
\beta_{\rm Tr}<0,
\qquad
\beta_{\rm M}<0,
\qquad
|\beta|\ll1.
\label{eq:common_ir_regime}
\end{equation}
This common regime constitutes the central architecture-level
observation: substantially different microscopic sequence dynamics
produce collective spectra that are both slow-mode rich and close to
the marginal flat-TDOS limit.

\begin{table*}[t]
\centering
\caption{
Comparison of the collective infrared dynamics measured in
Transformer and Mamba.
The infrared exponents denote representative resolved scaling
behavior and are not expected to be numerically identical between
architectures.
The corresponding long-time behavior follows from the TDOS--memory
relation established in the preceding sections.
}
\label{tab:transformer_mamba_comparison}

\begin{tabular*}{\textwidth}{@{\extracolsep{\fill}}lcc@{}}
\hline\hline
Property & Transformer & Mamba-370M \\
\hline
Microscopic mechanism
& Attention + MLP
& Selective SSM
\\
Temporal mechanism
& Context-dependent mixing
& Input-dependent state evolution
\\
Collective observable
& Full-block Jacobian TDOS
& Full-block Jacobian TDOS
\\
Block relaxation rate
& $-\ln|\zeta|$
& $-\ln|\zeta|$
\\
Infrared form
& $\rho(\lambda)\sim\lambda^\beta$
& $\rho(\lambda)\sim\lambda^\beta$
\\
Representative $\beta$
& $\sim-0.1$
& $\sim-0.17$
\\
IR character
& Nearly flat, weakly enhanced
& Nearly flat, weakly enhanced
\\
Long-time kernel
& $K(t)\sim t^{-0.9}$
& $K(t)\sim t^{-0.83}$
\\
Marginal limit
& Near $1/t$
& Near $1/t$
\\
\hline\hline
\end{tabular*}
\end{table*}

The dynamical consequence of this spectral organization follows from
the TDOS--memory relation derived in the preceding sections.
The representative exponents correspond to long-time kernels of
approximately $K(t)\sim t^{-0.9}$ for Transformer and
$K(t)\sim t^{-0.83}$ for Mamba.
Both systems therefore remain close to the marginal $1/t$ regime,
with the somewhat more negative Mamba exponent corresponding to a
modestly stronger long-time enhancement over the resolved scaling
range.

Taken together, these measurements reveal a separation between
microscopic implementation and macroscopic dynamical organization.
Transformer and Mamba construct their hidden-state transformations
through fundamentally different computational mechanisms, yet their
full-block dynamics independently develop a broad continuum of slow
collective modes and a nearly flat, weakly infrared-enhanced TDOS.
The present measurements do not establish identical spectra or a
universal numerical critical exponent across architectures.
They instead provide evidence for a more restricted and physically
meaningful form of architecture independence: distinct learned
sequence architectures can develop closely related infrared
collective organization.

The recurrence of this near-marginal regime in both attention-based
and selective state-space architectures suggests that infrared
slow-mode organization is not uniquely tied to either microscopic
mechanism.
Rather, it may represent a more general macroscopic dynamical regime
accessible to learned sequence models.

\section{V. Discussion}

The present results provide a direct view of how microscopic
relaxation dynamics is reorganized into macroscopic infrared
organization in a trained sequence model.
Mamba is particularly informative in this respect because its
state-space formulation exposes an explicit hierarchy of dynamical
scales that can be examined separately from the collective response
of the complete neural block.
The learned SSM generator contains a broad intrinsic distribution of
relaxation rates, the selective mechanism rescales these rates in an
input-dependent manner during inference, and the complete nonlinear
block develops a collective relaxation spectrum measured independently
through its Jacobian.

A central observation is that these dynamical levels are related but
not spectrally identical.
The intrinsic rates
\(
\lambda_A=e^{A_{\log}}
\)
define the temporal substrate encoded in the learned SSM, whereas
\(
\lambda_{\rm eff}(t)=\Delta_t\lambda_A
\)
describes its input-conditioned temporal rescaling.
The full-block rates,
\(
\lambda_{\rm block}=-\ln|\zeta|,
\)
characterize the collective response after gating, nonlinear
transformations, projections, channel mixing, and residual propagation
have been combined.
The full-block TDOS is therefore not a direct image of the spectrum
already present in the SSM generator.
Rather, the complete architecture substantially reorganizes the
microscopic relaxation hierarchy before the collective spectrum is
formed.

The selective mechanism provides a particularly transparent example
of this reorganization.
Through the input-dependent time step $\Delta_t$, the same intrinsic
SSM mode can acquire different effective relaxation times depending
on the processed input.
Small values of $\Delta_t$ shift the corresponding mode toward slower
effective relaxation, dynamically extending the temporal scales
available during inference.
The effective spectrum therefore describes an adaptive temporal
organization constructed from, but not identical to, the intrinsic
relaxation hierarchy.

The most significant scaling structure appears at the level of the
complete block.
As the retained sequence length increases, the full-block TDOS becomes
progressively smoother and better resolved toward the infrared, while
the cumulative exponent stabilizes near
$\beta_{\rm M}\simeq-0.17$ for the longest sequences examined.
This stabilization occurs while the number of modes sampled within
the same fitting interval increases substantially.
The observed behavior therefore reflects a reproducible statistical
organization of collective relaxation scales rather than simply the
appearance of additional eigenvalues in a larger Jacobian.

The sparsely populated extreme deep-infrared edge should nevertheless
be distinguished from this resolved scaling regime.
The smallest measured relaxation rates are more sensitive to
finite-size sampling and numerical resolution, and the infrared
exponent is therefore extracted from a common, well-resolved spectral
interval rather than from the minimum observed eigenvalues.
The central result is the emergence and stabilization of a broad
slow-mode continuum, not a particular numerical value of the smallest
relaxation rate.

The comparison with Transformer dynamics places this result in a
broader context.
Transformer and Mamba implement sequence processing through
fundamentally different microscopic mechanisms.
Mamba contains an explicit recurrent state-space evolution with
learned relaxation scales and input-dependent temporal selection,
whereas Transformer blocks organize hidden representations through
attention, nonlinear feed-forward transformations, and residual
propagation.
There is therefore no microscopic requirement that the two
architectures develop the same collective relaxation spectrum.

Nevertheless, when their complete hidden-state transformations are
analyzed using the same full-block Jacobian construction, both systems
exhibit a closely related infrared organization,
\begin{equation}
\rho(\lambda)
\sim
\lambda^\beta,
\qquad
\beta<0,
\qquad
|\beta|\ll1.
\label{eq:discussion_common_ir}
\end{equation}
The representative exponents,
$\beta_{\rm Tr}\sim-0.1$ and $\beta_{\rm M}\sim-0.17$, are not
numerically identical and should not be interpreted as a universal
critical exponent.
What is shared is the dynamical regime itself: a broad continuum of
collective relaxation modes with a nearly flat but weakly
infrared-enhanced TDOS.
The present measurements therefore do not establish universality in
the strong field-theoretic sense, but they provide evidence that
substantially different microscopic computational mechanisms can
produce closely related macroscopic relaxation organization.

Mamba also clarifies why a relaxation-spectrum description is more
than an alternative diagnostic representation of neural-network
dynamics.
In a Transformer, relaxation rates are inferred a posteriori from the
Jacobian spectrum of the learned hidden-state transformation, so the
architecture itself does not make it evident that relaxation
coordinates should provide a natural description of temporal
information processing.
In Mamba, by contrast, relaxation dynamics is an explicit part of the
computational mechanism.
The learned state-space generator directly encodes a hierarchy of
relaxation rates, and the selective mechanism dynamically rescales
these rates according to the processed input.
Relaxation scales are therefore not introduced solely by the present
spectral analysis; they are dynamical quantities through which
temporal information is propagated within the model.

At the same time, the collective TDOS cannot be reduced to this
explicit microscopic relaxation mechanism.
The progression
\[
\lambda_A
\longrightarrow
\lambda_{\rm eff}(x_t)
\longrightarrow
\{\lambda_\alpha^{\rm block}\}
\longrightarrow
\rho_{\rm block}(\lambda)
\]
shows that the learned intrinsic hierarchy is selectively rescaled and
then reorganized by the complete block before the collective spectrum
emerges.
The observation of a closely related collective infrared regime in
Transformers, where no corresponding SSM relaxation hierarchy is
explicitly built into the architecture, further indicates that the
full-block relaxation spectrum captures a level of dynamical
organization extending beyond the particular microscopic
implementation of Mamba.

The physical consequence of this collective organization is
long-memory dynamics.
Because the memory kernel is the Laplace transform of the relaxation
spectrum, a TDOS with
$\rho(\lambda)\sim\lambda^\beta$ produces
\begin{equation}
K(t)
\sim
t^{-(1+\beta)}
\label{eq:discussion_memory_scaling}
\end{equation}
over the corresponding resolved scaling regime.
An exactly flat TDOS gives the marginal $1/t$ form, while the weakly
negative exponents measured here imply slightly slower decay.
This prediction is examined directly in Appendix~D, where the memory
kernel is constructed from the measured full-block relaxation modes.
The full-spectrum kernel remains long lived as progressively slower
modes are resolved, while the kernel restricted to the common
infrared fitting window exhibits the temporal scaling expected from
the measured TDOS exponent.
For Mamba, $\beta_{\rm M}\simeq-0.17$ corresponds to
$K(t)\sim t^{-0.83}$ over the associated temporal regime.

This result also distinguishes microscopic state-space memory from
collective memory.
In Mamba, temporal persistence is explicitly implemented through the
propagation of an internal SSM state.
The collective memory characterized by the full-block TDOS has a
different status: it describes the relaxation organization of the
complete learned hidden-state transformation.
Mamba therefore allows microscopic memory implementation and
macroscopic collective memory to be observed separately within the
same architecture.
The Transformer comparison strengthens this distinction, since a
closely related near-marginal collective spectrum appears without the
same explicit recurrent SSM state or selective temporal rescaling.

Within Cognitive Field Theory, these observations have a natural
interpretation.
Long-lived collective memory is associated not with a single
microscopic memory variable but with the infrared organization of a
continuum of collective relaxation modes.
The collective TDOS $\rho(\lambda)$ determines the time-domain memory
kernel $K(t)$ and the corresponding frequency-domain self-energy
$\Sigma_R(\omega)$, connecting infrared spectral organization to
non-Markovian memory feedback.
Mamba provides a particularly transparent test of this distinction
because its microscopic relaxation substrate can be measured directly
and compared with the independently reconstructed collective spectrum
of the complete block.

The results therefore suggest a separation between
architecture-specific computation and macroscopic dynamical
organization.
Microscopic mechanisms determine how each architecture represents and
propagates information, whereas the collective TDOS characterizes how
the resulting high-dimensional dynamics is organized over relaxation
time scales.
The observation of closely related infrared organization in
Transformer and Mamba suggests that the macroscopic relaxation
structure can be substantially less architecture dependent than the
microscopic mechanisms from which it emerges.

Several limitations define the scope of this conclusion.
The present Mamba measurements focus primarily on Mamba-370M, and the
cross-architecture comparison involves two major classes of sequence
models rather than an exhaustive set of neural architectures.
The measured exponents characterize finite, experimentally resolved
infrared intervals and should not be extrapolated directly to the
exact $\lambda\rightarrow0$ limit.
Most importantly, the present measurements characterize trained
models and do not determine how the intrinsic, effective, and
collective relaxation spectra develop throughout optimization.

These limitations define direct extensions of the present work.
Measurements across additional model scales and independently trained
instances would further test the robustness of the observed infrared
regime, while comparisons with recurrent networks and other
state-space architectures would determine its architectural range.
Training checkpoints would be particularly informative because they
would allow the intrinsic SSM spectrum, the input-conditioned
effective spectrum, and the full-block collective TDOS to be followed
simultaneously during learning.
Such measurements could determine how microscopic temporal structure
is progressively reorganized into collective infrared dynamics and,
through the corresponding memory kernel and self-energy, whether the
collective response approaches the near-critical regime observed in
Transformer dynamics.

Taken together, the present results identify an experimentally
accessible form of cross-architecture collective organization.
Two substantially different sequence architectures exhibit a broad
collective slow-mode continuum, a nearly flat but weakly
infrared-enhanced TDOS, and the corresponding near-marginal
long-memory dynamics.
Mamba further demonstrates that relaxation dynamics can be an explicit
component of learned temporal computation while the resulting
collective TDOS remains distinct from, and cannot be reduced to, the
microscopic SSM relaxation hierarchy.
The emergence of closely related collective infrared organization in
Mamba and Transformer therefore suggests that near-marginal slow-mode
dynamics may represent a more general macroscopic principle of learned
sequence processing rather than a property tied to a particular
microscopic memory mechanism.

\section{VI. Conclusion}

We have investigated the infrared organization of learned dynamics in
Mamba and compared it with the collective relaxation structure
observed in Transformers.

The explicit state-space formulation of Mamba makes it possible to
resolve distinct dynamical levels: the intrinsic relaxation spectrum
encoded in the learned SSM generator, its input-dependent temporal
rescaling through the selective mechanism, and the collective
relaxation spectrum of the complete nonlinear block.

The measurements show that these levels are related but not
spectrally identical.
Mamba contains a broad intrinsic hierarchy of relaxation scales, which
is dynamically rescaled by the input-dependent selective time step.
After the remaining nonlinear transformations, gating, projections,
and residual propagation are included, the complete block develops a
reorganized collective spectrum containing a broad continuum of slow
relaxation modes.

With increasing sequence length, this infrared continuum becomes
progressively better resolved while its cumulative scaling stabilizes.
For the longest sequences examined, the resolved Mamba TDOS is well
described by $\rho(\lambda)\sim\lambda^\beta$ with
$\beta\simeq-0.17$, corresponding to a nearly flat but weakly
infrared-enhanced spectrum.

A central result is that a closely related collective organization is
observed in Transformers despite their fundamentally different
microscopic dynamics.
Using the same full-block Jacobian construction, Transformer and Mamba
both exhibit broad near-flat infrared spectra with small negative
exponents, of order $\beta_{\rm Tr}\sim-0.1$ and
$\beta_{\rm M}\sim-0.17$, respectively.
The numerical values are not identical and should not be interpreted
as a universal critical exponent.
Rather, the common observation is a slow-mode-rich collective regime
characterized by $\beta<0$ and $|\beta|\ll1$.

This spectral organization produces near-marginal long-memory
dynamics, since $\rho(\lambda)\sim\lambda^\beta$ corresponds to
$K(t)\sim t^{-(1+\beta)}$ over the associated scaling regime, as also
confirmed directly from the measured relaxation spectrum.
Thus, two architectures that implement sequence processing through
substantially different microscopic mechanisms develop collective
dynamics close to the marginal $1/t$ memory form.

Within Cognitive Field Theory, these results support a separation
between microscopic memory mechanisms and macroscopic infrared
organization.
Mamba makes this distinction particularly transparent because its
internal relaxation hierarchy can be observed directly, whereas the
Transformer comparison shows that a closely related collective
infrared regime does not require an explicit state-space mechanism.

The present measurements therefore provide evidence that
near-marginal slow-mode organization may represent a more general
macroscopic dynamical principle of learned sequence processing rather
than a property tied to a particular neural architecture.

\vspace{6pt}
\emph{Acknowledgements}---This work was partially supported by the Institute of Information \& Communications Technology Planning \& Evaluation (IITP) grant 
funded by the Korea government (MSIT) (IITP-RS-2025-02214780).

The author acknowledges the support of ChatGPT (GPT-5, OpenAI) for assistance in literature review and conceptual structuring during early development.

\clearpage
\appendix

\renewcommand{\thefigure}{A\arabic{figure}}
\renewcommand{\theequation}{A\arabic{equation}}

\setcounter{figure}{0}
\setcounter{equation}{0}

\vspace*{1.5cm}
{\centering\large\bfseries Supplementary Materials\par}
\vspace{1.0cm}

\section{Appendix A: Causal Jacobian structure and exact token-resolved construction of the full relaxation spectrum}

\label{app:causal_jacobian_tdos}

In this appendix we establish the relation between the token-resolved
Jacobian spectra used in the numerical analysis and the eigenvalue
spectrum of the full vectorized sequence mapping.

The key observation is that a strictly causal sequence model generates
a block-lower-triangular Jacobian in token space.  The characteristic
polynomial of such a matrix factorizes exactly into those of its
token-diagonal blocks.  Consequently, the eigenvalue spectrum of the
full sequence Jacobian is the union of the spectra obtained from the
token-diagonal Jacobians.

This result provides the mathematical basis for constructing the full
relaxation spectrum and the corresponding time-scale density of states
(TDOS) without explicitly forming the full sequence Jacobian.  The
result concerns the eigenvalue spectrum itself; off-diagonal token
couplings remain essential for propagation, eigenvector structure,
non-normality, and transient dynamics.

\vspace{10pt}

\noindent
\emph{(I) Full sequence Jacobian and token-resolved blocks.}

Consider a sequence of length \(T\),
\begin{equation}
H
=
(h_1,h_2,\ldots,h_T),
\qquad
h_t\in\mathbb{R}^{d},
\end{equation}
where \(d\) denotes the hidden-state dimension of the analyzed
representation.

A neural block defines a sequence-to-sequence mapping
\begin{equation}
H'
=
F(H).
\end{equation}
After vectorizing the sequence,
\begin{equation}
\operatorname{vec}(H)
\in
\mathbb{R}^{Td},
\end{equation}
the full Jacobian is
\begin{equation}
J_{\mathrm{full}}
=
\frac{
\partial\operatorname{vec}(H')
}{
\partial\operatorname{vec}(H)
}
\in
\mathbb{R}^{Td\times Td}.
\label{eq:app_full_jacobian}
\end{equation}

It is convenient to decompose this matrix into \(d\times d\)
token-resolved blocks,
\begin{equation}
J_{\mathrm{full}}
=
\begin{pmatrix}
J_{11} & J_{12} & \cdots & J_{1T}\\
J_{21} & J_{22} & \cdots & J_{2T}\\
\vdots & \vdots & \ddots & \vdots\\
J_{T1} & J_{T2} & \cdots & J_{TT}
\end{pmatrix},
\label{eq:app_full_block_jacobian}
\end{equation}
where
\begin{equation}
J_{ts}
=
\frac{\partial h'_t}{\partial h_s}.
\label{eq:app_token_block}
\end{equation}
The diagonal block
\begin{equation}
J_{tt}
=
\frac{\partial h'_t}{\partial h_t}
\label{eq:app_token_diag_jacobian}
\end{equation}
describes the local linear response of output token \(t\) to a
perturbation of the corresponding input token.

By contrast, an off-diagonal block
\begin{equation}
J_{ts},
\qquad
t\neq s,
\end{equation}
describes the propagation of a perturbation between different token
positions.

\vspace{10pt}

\noindent
\emph{(II) Causality and block-triangular structure.}

For a strictly causal sequence mapping, the output at position \(t\)
can depend only on the present and preceding token representations,
\begin{equation}
h'_t
=
F_t(h_1,h_2,\ldots,h_t).
\label{eq:app_causal_mapping}
\end{equation}
It therefore follows directly that
\begin{equation}
\frac{\partial h'_t}{\partial h_s}
=
0,
\qquad
s>t.
\label{eq:app_future_derivative_zero}
\end{equation}
Hence the full sequence Jacobian takes the block-lower-triangular form
\begin{equation}
J_{\mathrm{full}}
=
\begin{pmatrix}
J_{11} & 0      & 0      & \cdots & 0\\
J_{21} & J_{22} & 0      & \cdots & 0\\
J_{31} & J_{32} & J_{33} & \cdots & 0\\
\vdots & \vdots & \vdots & \ddots & \vdots\\
J_{T1} & J_{T2} & J_{T3} & \cdots & J_{TT}
\end{pmatrix}.
\label{eq:app_causal_triangular}
\end{equation}

This structure does not imply that different token positions are
dynamically independent.  In a causal state-space model, the blocks
\begin{equation}
J_{ts}\neq0,
\qquad
s<t,
\end{equation}
encode the propagation of information from earlier positions to later
positions through the recurrent state-space dynamics.

Causality imposes only the absence of the reverse dependence on future
positions,
\begin{equation}
J_{ts}=0,
\qquad
s>t.
\end{equation}

The resulting triangular structure is sufficient to determine the
eigenvalue spectrum exactly.

\vspace{10pt}

\noindent
\emph{(III) Exact factorization of the full eigenvalue spectrum.}

Let \(\mu\) denote an eigenvalue of the full sequence Jacobian.
Its characteristic polynomial is
\begin{equation}
P(\mu)
=
\det
\left(
\mu I_{Td}
-
J_{\mathrm{full}}
\right).
\label{eq:app_characteristic_full}
\end{equation}
Since \(J_{\mathrm{full}}\) is block lower triangular,
\(\mu I_{Td}-J_{\mathrm{full}}\) is also block lower triangular,
\begin{equation}
\mu I_{Td}
-
J_{\mathrm{full}}
=
\begin{pmatrix}
\mu I_d-J_{11} & 0 & \cdots & 0\\
-J_{21} & \mu I_d-J_{22} & \cdots & 0\\
\vdots & \vdots & \ddots & \vdots\\
-J_{T1} & -J_{T2} & \cdots & \mu I_d-J_{TT}
\end{pmatrix}.
\end{equation}

The determinant of a block-triangular matrix is the product of the
determinants of its diagonal blocks.  Therefore,
\begin{equation}
P(\mu)
=
\prod_{t=1}^{T}
\det
\left(
\mu I_d-J_{tt}
\right).
\label{eq:app_characteristic_factorization}
\end{equation}

The roots of the full characteristic polynomial are consequently
exactly the roots of the individual token-block characteristic
polynomials.
Thus,
\begin{equation}
\sigma(J_{\mathrm{full}})
=
\bigcup_{t=1}^{T}
\sigma(J_{tt}),
\label{eq:app_full_spectrum_union}
\end{equation}
where the union includes algebraic multiplicities.

Importantly, Eq.~(\ref{eq:app_full_spectrum_union}) does not require
the off-diagonal blocks to be small.  The causal couplings
\begin{equation}
J_{ts},
\qquad
s<t,
\end{equation}
may be large while the eigenvalue identity remains exact.

The reduction therefore follows from causality and triangularity,
rather than from a weak-coupling or local-token approximation.

The same spectral reduction applies to causal Transformer
architectures.  Although the microscopic mechanisms of sequence
propagation are different, causal masking in an autoregressive
Transformer likewise prevents the representation at position $t$
from depending on future token positions.  Its sequence Jacobian is
therefore block lower triangular in token space, and its full
eigenvalue spectrum is again determined exactly by the collection of
token-diagonal spectra.  Thus, the spectral identity derived here is
not specific to the state-space implementation of Mamba, but follows
more generally from strict causal sequence processing.

\vspace{10pt}

\noindent
\emph{(IV) Relaxation rates and token-resolved construction of the TDOS.}

Let the eigenvalues of a token-diagonal block satisfy
\begin{equation}
J_{tt}u_{t\alpha}
=
\mu_{t\alpha}u_{t\alpha},
\qquad
\alpha=1,\ldots,d.
\label{eq:app_token_eigenproblem}
\end{equation}
Because the analyzed neural block defines a discrete mapping, repeated
application of a mode gives
\begin{equation}
u_{t\alpha}(n)
=
\mu_{t\alpha}^{\,n}
u_{t\alpha}(0).
\end{equation}
Writing
\begin{equation}
\mu_{t\alpha}
=
|\mu_{t\alpha}|
e^{i\theta_{t\alpha}},
\end{equation}
gives
\begin{equation}
u_{t\alpha}(n)
=
|\mu_{t\alpha}|^n
e^{in\theta_{t\alpha}}
u_{t\alpha}(0).
\end{equation}

The magnitude therefore controls contraction or expansion, while the
phase controls rotation in the corresponding eigenspace.
For the stable sector,
\begin{equation}
0<|\mu_{t\alpha}|<1,
\end{equation}
we define
\begin{equation}
|\mu_{t\alpha}|
=
e^{-\lambda_{t\alpha}},
\end{equation}
so that
\begin{equation}
\lambda_{t\alpha}
=
-\log|\mu_{t\alpha}|.
\label{eq:app_discrete_relaxation_rate}
\end{equation}

The limit
\begin{equation}
|\mu_{t\alpha}|
\rightarrow1^{-}
\end{equation}
corresponds to
\begin{equation}
\lambda_{t\alpha}
\rightarrow0^{+},
\end{equation}
and therefore identifies the slow infrared sector of the relaxation
spectrum.

The token-resolved empirical TDOS may be written as
\begin{equation}
\rho_t(\lambda)
=
\frac{1}{d}
\sum_{\alpha=1}^{d}
\delta
\left(
\lambda-\lambda_{t\alpha}
\right),
\label{eq:app_token_resolved_tdos}
\end{equation}
with the normalization modified accordingly when only stable modes are
retained.

Using Eq.~(\ref{eq:app_full_spectrum_union}), the full sequence TDOS is
\begin{align}
\rho_T(\lambda)
&=
\frac{1}{Td}
\sum_{t=1}^{T}
\sum_{\alpha=1}^{d}
\delta
\left(
\lambda-\lambda_{t\alpha}
\right)
\nonumber\\
&=
\frac{1}{T}
\sum_{t=1}^{T}
\rho_t(\lambda).
\label{eq:app_full_tdos_token_average}
\end{align}
Thus, for a strictly causal mapping, the eigenvalue spectrum of the
full sequence Jacobian is determined exactly by the collection of
token-diagonal spectra.

The token-resolved computation used in the numerical analysis therefore
does not approximate the eigenvalue TDOS of the full causal sequence
Jacobian.  It provides an exact decomposition of that spectrum into
computationally accessible \(d\times d\) blocks.

\vspace{10pt}

\noindent
\emph{(V) Computational reduction.}

The full vectorized sequence Jacobian has dimension
\begin{equation}
Td\times Td
\end{equation}
and therefore contains
\begin{equation}
T^2d^2
\end{equation}
matrix elements.
Explicit construction and diagonalization of this matrix rapidly
become computationally prohibitive for realistic sequence lengths and
hidden dimensions.

By contrast, the causal spectral identity
(\ref{eq:app_full_spectrum_union}) permits the calculation to be
performed using the \(T\) token-diagonal matrices
\begin{equation}
J_{11},J_{22},\ldots,J_{TT},
\end{equation}
each of dimension
\begin{equation}
d\times d.
\end{equation}

The total number of explicitly stored diagonal-block elements therefore
scales as
\begin{equation}
Td^2,
\end{equation}
rather than
\begin{equation}
T^2d^2.
\end{equation}
This reduction makes direct eigenspectral measurements possible while
preserving the exact eigenvalue spectrum relevant to the TDOS.

It should be emphasized that this is a reduction of the
\emph{eigenvalue problem}, not a reconstruction of the complete
Jacobian matrix.

\vspace{10pt}

\noindent
\emph{(VI) Information retained and discarded by the TDOS.}

Although the off-diagonal causal blocks do not alter the eigenvalue
set of a block-triangular Jacobian, they remain dynamically important.

A simple two-dimensional example illustrates this distinction.
Consider
\begin{equation}
J
=
\begin{pmatrix}
a & 0\\
b & a
\end{pmatrix}.
\label{eq:app_nonnormal_example}
\end{equation}
The eigenvalues are
\begin{equation}
\mu_1=\mu_2=a,
\end{equation}
independent of \(b\).

However, repeated application gives
\begin{equation}
J^n
=
\begin{pmatrix}
a^n & 0\\
n b a^{n-1} & a^n
\end{pmatrix}.
\label{eq:app_nonnormal_power}
\end{equation}
Thus the off-diagonal coupling \(b\) can generate substantial transient
propagation even though it does not modify the eigenvalues.

More generally, the off-diagonal token blocks remain dynamically
important because they can affect token-to-token propagation,
eigenvector geometry, non-normality, and transient amplification.

The TDOS should therefore be interpreted specifically as a density of
relaxation scales,
\begin{equation}
\rho(\lambda),
\end{equation}
rather than as a complete description of all aspects of sequence
dynamics.

This distinction is particularly relevant for causal state-space
models.  Their recurrent state dynamics can produce strong
past-to-future propagation through the off-diagonal blocks of
\(J_{\mathrm{full}}\), while the relaxation-rate spectrum remains
exactly determined by the token-diagonal blocks.

\vspace{10pt}

\noindent
\emph{(VII) Relation to the macroscopic TDOS.}

The preceding construction establishes the spectral hierarchy
\begin{equation}
J_{\mathrm{full}}
\longrightarrow
\{J_{tt}\}
\longrightarrow
\{\mu_{t\alpha}\}
\longrightarrow
\{\lambda_{t\alpha}\}
\longrightarrow
\rho_T(\lambda).
\label{eq:app_jacobian_tdos_hierarchy}
\end{equation}

The first equality underlying this hierarchy is exact for a strictly
causal mapping: the eigenvalue spectrum of \(J_{\mathrm{full}}\) is
completely determined by the token-diagonal spectra.

A separate question is whether the finite-sequence empirical density
\begin{equation}
\rho_T(\lambda)
\end{equation}
converges toward a reproducible macroscopic spectral envelope as the
number of sampled token positions increases.

That question concerns statistical self-averaging rather than the
linear-algebraic spectral identity derived here.  As shown in
Appendix~B, if token-resolved spectra sample a common learned spectral
organization, one may write
\begin{equation}
\rho_t(\lambda)
=
\rho_{\mathrm{bulk}}(\lambda)
+
\delta\rho_t(\lambda),
\end{equation}
so that increasing sequence length suppresses finite-mode fluctuations
and progressively resolves the underlying macroscopic TDOS.

The two results therefore have distinct but complementary roles:
causal triangularity provides the exact spectral decomposition of
the full sequence Jacobian into token-diagonal spectra, whereas
spectral self-averaging explains the emergence of a macroscopic TDOS
from these token-resolved spectral measures.

Together they provide the mathematical basis for interpreting the
token-resolved Jacobian measurements used in the main text as a
direct probe of the collective relaxation spectrum of the causal
sequence model.

\clearpage
\appendix

\renewcommand{\thefigure}{B\arabic{figure}}
\renewcommand{\theequation}{B\arabic{equation}}

\setcounter{figure}{0}
\setcounter{equation}{0}

\section{Appendix B: Token-resolved spectral self-averaging and the macroscopic TDOS}

\label{app:token_tdos_selfaveraging}

In this appendix we clarify the relation between the token-resolved
relaxation spectra and the full time-scale density of states
used in the main text.

The central point is that, for a causal sequence mapping, the full
Jacobian has a block-triangular structure.  Its eigenvalue spectrum is
therefore determined exactly by the spectra of the token-diagonal
Jacobian blocks.  The full TDOS can consequently be interpreted as an
average over token-resolved empirical spectral measures.

This observation also provides a statistical interpretation of the
sequence-length dependence discussed in the main text.  If different
token positions sample a common learned spectral organization, an
individual token already probes the same underlying TDOS, but with
finite-mode fluctuations.  Increasing the sequence length then
improves the statistical resolution of this spectrum rather than
generating the spectral envelope itself.

\vspace{10pt}

\noindent
\emph{(I) Token-resolved and full relaxation spectra.}

Consider a causal neural sequence mapping acting on
\begin{equation}
H=(h_1,h_2,\ldots,h_T),
\qquad
h_t\in\mathbb{R}^{d}.
\end{equation}
The token-diagonal Jacobian block is
\begin{equation}
J_{tt}
=
\frac{\partial h'_t}{\partial h_t},
\end{equation}
with eigenvalues
\begin{equation}
J_{tt}u_{t\alpha}
=
\mu_{t\alpha}u_{t\alpha}.
\end{equation}
For stable modes,
\(0<|\mu_{t\alpha}|<1\), we define the discrete relaxation rate as
\begin{equation}
\lambda_{t\alpha}
=
-\log|\mu_{t\alpha}|.
\label{eq:app_token_lambda}
\end{equation}
The corresponding token-resolved empirical TDOS is
\begin{equation}
\rho_t(\lambda)
=
\frac{1}{d}
\sum_{\alpha=1}^{d}
\delta
\left(
\lambda-\lambda_{t\alpha}
\right),
\label{eq:app_token_tdos}
\end{equation}
where the normalization may equivalently be restricted to the number
of stable modes when only the stable sector is retained.

For a strictly causal sequence mapping, the full Jacobian has the
block-lower-triangular form
\begin{equation}
J_{\mathrm{full}}
=
\begin{pmatrix}
J_{11} & 0 & 0 & \cdots \\
J_{21} & J_{22} & 0 & \cdots \\
J_{31} & J_{32} & J_{33} & \cdots \\
\vdots & \vdots & \vdots & \ddots
\end{pmatrix}.
\end{equation}

The spectrum of a block-triangular matrix is the union of the spectra
of its diagonal blocks,
\begin{equation}
\sigma(J_{\mathrm{full}})
=
\bigcup_{t=1}^{T}
\sigma(J_{tt}).
\label{eq:app_full_union}
\end{equation}

Consequently, pooling the relaxation rates over all token positions
gives
\begin{align}
\rho_T(\lambda)
&=
\frac{1}{Td}
\sum_{t=1}^{T}
\sum_{\alpha=1}^{d}
\delta
\left(
\lambda-\lambda_{t\alpha}
\right)
\nonumber\\
&=
\frac{1}{T}
\sum_{t=1}^{T}
\rho_t(\lambda).
\label{eq:app_full_tdos_average}
\end{align}

Thus the pooled TDOS is not merely a numerical averaging procedure.
Under causality, it is the empirical relaxation-rate density associated
with the full vectorized sequence Jacobian.

\vspace{10pt}

\noindent
\emph{(II) Common spectral envelope and self-averaging.}

Suppose that token-resolved spectra probe a common learned spectral
distribution \(\rho(\lambda)\).  The spectrum at an individual token
may then be written schematically as
\begin{equation}
\rho_t(\lambda)
=
\rho(\lambda)
+
\delta\rho_t(\lambda),
\label{eq:app_token_fluctuation}
\end{equation}
where \(\delta\rho_t(\lambda)\) denotes finite-mode and
context-dependent fluctuations.

For a stationary spectral ensemble,
\begin{equation}
\mathbb{E}
\left[
\rho_t(\lambda)
\right]
=
\rho(\lambda),
\end{equation}
and Eq.~(\ref{eq:app_full_tdos_average}) becomes
\begin{equation}
\rho_T(\lambda)
=
\rho(\lambda)
+
\frac{1}{T}
\sum_{t=1}^{T}
\delta\rho_t(\lambda).
\label{eq:app_tdos_selfaverage}
\end{equation}

If the token-resolved fluctuations are independent or sufficiently
weakly correlated, their contribution to the sequence average is
suppressed with increasing \(T\).  In the ideal independent limit,
\begin{equation}
\mathrm{Var}
\left[
\rho_T(\lambda)
\right]
\propto
\frac{1}{T}.
\end{equation}

More generally, correlations between neighboring token spectra replace
\(T\) by an effective number of statistically independent samples
\(T_{\mathrm{eff}}\).  Provided that the spectral correlations remain
sufficiently short ranged, one obtains the self-averaging limit
\begin{equation}
\rho_T(\lambda)
\longrightarrow
\rho(\lambda)
\qquad
(T\rightarrow\infty).
\label{eq:app_tdos_bulk_limit}
\end{equation}

Accordingly, the qualitative TDOS envelope may already be visible at
the level of individual tokens, while longer sequences suppress
finite-sample fluctuations and reveal the common spectral organization
with progressively higher resolution.

\vspace{10pt}

\noindent
\emph{(III) Finite-sampling fluctuations in the infrared.}

The enhanced fluctuations observed in the deepest infrared have a
simple statistical origin.
Consider a spectral bin \(B\) with probability mass
\begin{equation}
p_B
=
\int_B
\rho(\lambda)\,d\lambda.
\end{equation}

If \(N\) relaxation modes effectively sample this distribution, the
expected number of modes in the bin is
\begin{equation}
\langle n_B\rangle
=
Np_B.
\end{equation}

Neglecting correlations for the purpose of estimating the sampling
scale, binomial counting gives
\begin{equation}
\frac{\delta n_B}{n_B}
\sim
\frac{1}{\sqrt{Np_B}}
\qquad
(p_B\ll1).
\label{eq:app_bin_fluctuation}
\end{equation}

For \(T\) tokens with \(d\) modes per token, the nominal number of
sampled modes is \(N\simeq Td\).  If the infrared TDOS follows
\begin{equation}
\rho(\lambda)
\simeq
C_\beta\lambda^\beta,
\label{eq:app_ir_tdos_power}
\end{equation}
and logarithmic bins are used, then
\(\Delta\lambda\propto\lambda\), so that
\begin{equation}
p_B
\simeq
\rho(\lambda)\Delta\lambda
\propto
\lambda^{1+\beta}.
\end{equation}

Equation~(\ref{eq:app_bin_fluctuation}) therefore gives the approximate
finite-sampling scaling
\begin{equation}
\frac{\delta\rho}{\rho}
\sim
\frac{1}{\sqrt{Td}}\,
\lambda^{-(1+\beta)/2}.
\label{eq:app_ir_sampling_fluctuation}
\end{equation}
For the nearly flat infrared spectrum observed in the main text,
\begin{equation}
\beta\simeq0,
\end{equation}
this reduces to
\begin{equation}
\frac{\delta\rho}{\rho}
\sim
\frac{1}{\sqrt{Td\,\lambda}}.
\label{eq:app_flat_tdos_fluctuation}
\end{equation}

Thus even a genuinely flat underlying TDOS is expected to exhibit
increasing relative fluctuations as \(\lambda\rightarrow0\) when
estimated from a finite number of modes.  Strong fluctuations in the
extreme infrared therefore do not by themselves imply a breakdown of
the underlying spectral envelope.

The same counting argument provides an estimate of the smallest
relaxation scale that can be statistically resolved.  Requiring at
least order-one occupation of a logarithmic bin gives
\begin{equation}
Td\,C_\beta
\lambda_{\mathrm{res}}^{1+\beta}
\sim
1,
\end{equation}
and hence
\begin{equation}
\lambda_{\mathrm{res}}
\sim
(Td)^{-1/(1+\beta)}.
\label{eq:app_lambda_resolution}
\end{equation}
For \(\beta\simeq0\),
\begin{equation}
\lambda_{\mathrm{res}}
\sim
\frac{1}{Td}.
\end{equation}

These relations should be understood as finite-sampling estimates:
correlations among eigenmodes and among neighboring token spectra
reduce the effective number of statistically independent samples.

\vspace{10pt}

\noindent
\emph{(IV) Interpretation of the macroscopic TDOS.}

The preceding results distinguish two conceptually different effects
of increasing sequence length.

First, the causal Jacobian identity
(\ref{eq:app_full_union}) shows that increasing \(T\) adds additional
token-resolved relaxation spectra to the full sequence spectrum.
Second, if these spectra share a common learned envelope, the enlarged
sequence provides progressively better statistical sampling of that
same spectral organization.

The resulting hierarchy is therefore
\begin{equation}
J_{tt}
\longrightarrow
\{\lambda_{t\alpha}\}
\longrightarrow
\rho_t(\lambda)
\longrightarrow
\rho_T(\lambda)
\longrightarrow
\rho_{\mathrm{bulk}}(\lambda).
\label{eq:app_tdos_hierarchy}
\end{equation}

In this interpretation, increasing sequence length does not by itself
create the infrared TDOS.  Rather, the token-resolved Jacobian spectra
already contain the local spectral tendency, while pooling over tokens
suppresses finite-mode fluctuations and extends the reliably resolved
window toward smaller relaxation rates.

This provides a statistical basis for treating the TDOS as a
macroscopic dynamical observable of the learned sequence model.
Individual Jacobian eigenvalues and individual token spectra remain
microscopic and context dependent, whereas their empirical spectral
density can converge toward a reproducible large-scale form.
Accordingly, the persistence of the same infrared TDOS envelope across
token positions and sequence lengths constitutes a direct test of
whether the observed spectral structure reflects a learned collective
organization rather than an artifact of finite sequence pooling.

\clearpage
\appendix

\renewcommand{\thefigure}{C\arabic{figure}}
\renewcommand{\theequation}{C\arabic{equation}}

\setcounter{figure}{0}
\setcounter{equation}{0}

\section{Appendix C: FP64 numerical validation of the full-block Jacobian TDOS}
\label{app:fp64_validation}

The full-block Jacobian spectrum measured in the main text extends
into an extremely slow-relaxation regime, where the inferred
relaxation rates become very small.
Because the pretrained Mamba checkpoint is originally stored at finite
numerical precision and some operations in the standard implementation
may be evaluated in FP32, it is important to determine whether the
observed deep-infrared structure is sensitive to the numerical
precision used in the Jacobian calculation.
We therefore repeated the full-block spectral analysis using an FP64
numerical implementation and compared the resulting spectra directly
with the corresponding FP32 measurement.

The purpose of this calculation is not to assign FP64 information to
the pretrained parameters themselves.
Promoting parameters originally stored at lower precision to
double precision does not recover information absent from the
checkpoint.
Rather, the FP64 calculation reduces subsequent numerical errors in
the evaluation of the nonlinear block transformation, automatic
differentiation, Jacobian construction, and eigenspectral analysis.
It therefore provides a direct numerical robustness test of the
collective relaxation spectrum extracted from the trained model.

For the complete hidden-state mapping
\begin{equation}
H_{\ell+1}
=
F_\ell(H_\ell),
\end{equation}
we construct the full-block Jacobian
\begin{equation}
J_\ell
=
\frac{
\partial\,{\rm vec}(H_{\ell+1})
}{
\partial\,{\rm vec}(H_\ell)
},
\label{eq:fp64_jacobian}
\end{equation}
and obtain its eigenspectrum from
\begin{equation}
J_\ell u_\alpha
=
\zeta_\alpha u_\alpha.
\end{equation}
As in the main analysis, the relaxation rate associated with each
stable Jacobian eigenmode is defined by
\begin{equation}
\lambda_\alpha
=
-\ln|\zeta_\alpha|,
\label{eq:fp64_relaxation_rate}
\end{equation}
and the corresponding full-block time-scale density of states is
\begin{equation}
\rho_{\rm block}(\lambda)
=
\frac{1}{N}
\sum_\alpha
\delta
\left(
\lambda-\lambda_\alpha
\right).
\label{eq:fp64_tdos}
\end{equation}

\begin{figure}[t]
\centering
\includegraphics[scale=0.65, trim= 0.3cm 0.5cm 0cm 0cm]{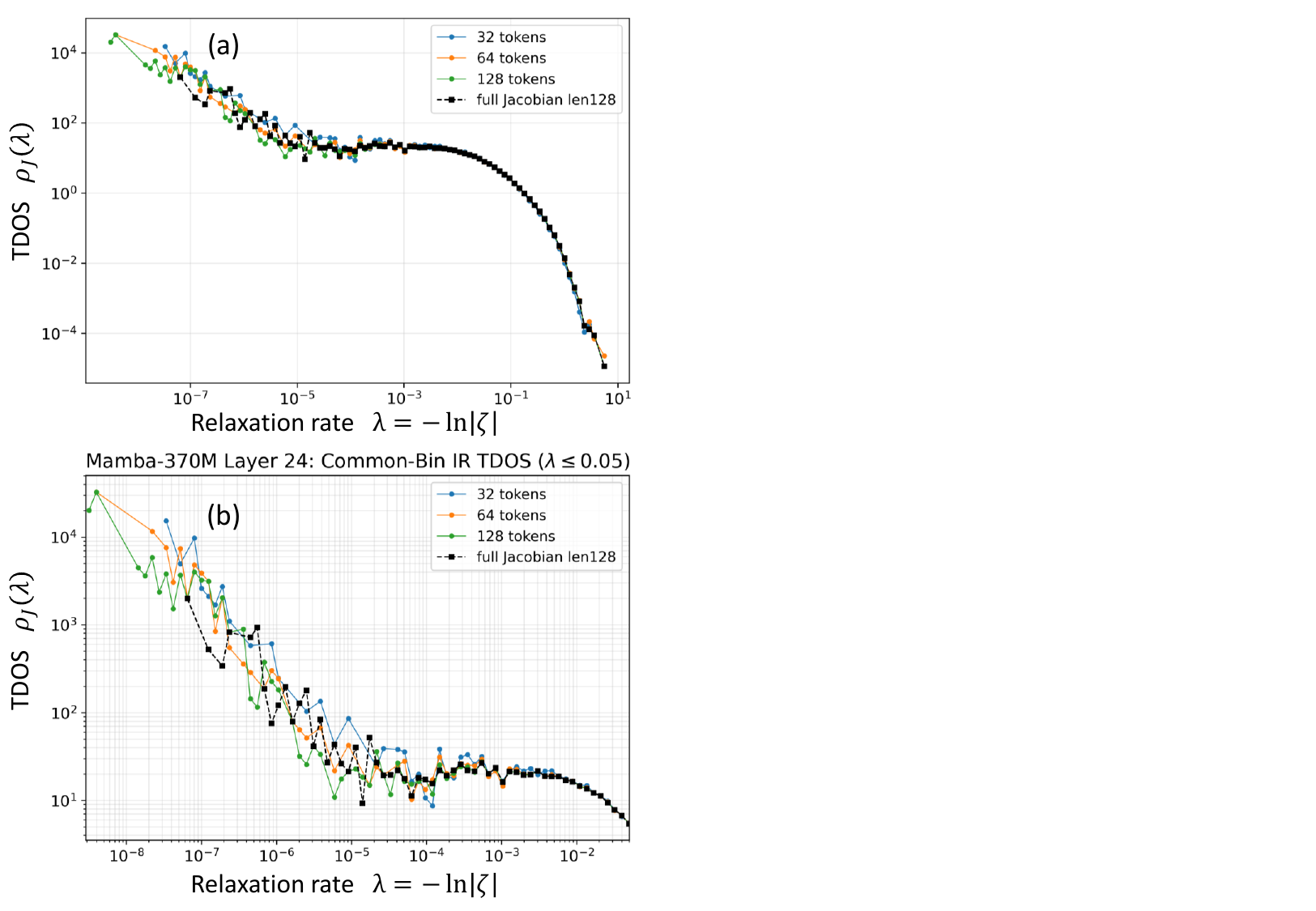}
\caption{
FP64 numerical validation of the Mamba-370M full-block
Jacobian TDOS.
(a) Full-block time-scale density of states at layer~24 obtained from
FP64 Jacobian calculations for retained sequence lengths
$L=32$, $64$, and $128$.
The corresponding FP32 spectrum for $L=128$ is overlaid as a black
dashed curve with square markers.
The FP32 and FP64 measurements recover essentially the same collective
spectral envelope throughout the statistically well-resolved regime,
including the broad slow-relaxation sector, the intermediate spectral
region, and the ultraviolet rolloff.
(b) The same spectra restricted to the infrared region
$\lambda\leq0.05$ using common logarithmic bins.
The agreement between FP32 and FP64 persists throughout the resolved
infrared continuum, while larger bin-to-bin fluctuations appear in
the sparsely populated extreme-infrared tail.
The FP64 measurements further resolve the collective spectrum toward
relaxation rates of order $\lambda\sim10^{-8}$.
The persistence of the same large-scale spectral organization under
increased numerical precision demonstrates that the observed
collective infrared structure is not an artifact of the arithmetic
precision used in the full-block Jacobian calculation.
}
\label{fig:fp64_fullblock_tdos}
\end{figure}

Figure~\ref{fig:fp64_fullblock_tdos}(a) shows the resulting FP64
full-block TDOS of Mamba-370M at layer~24 for retained sequence
lengths $L=32$, $64$, and $128$.
For direct comparison, the corresponding FP32 spectrum at
$L=128$ is overlaid on the same axes.
The FP64 spectra for the three sequence lengths reproduce essentially
the same large-scale spectral envelope.
More importantly, the FP32 and FP64 measurements also closely overlap
throughout the statistically well-resolved part of the spectrum,
extending from the slow-relaxation sector through the intermediate
regime and into the rapidly decaying ultraviolet tail.

This direct overlap provides a stronger numerical test than the FP64
measurement alone.
Changing the arithmetic precision modifies the numerical evaluation
of the block response and its eigenspectrum, but does not qualitatively
reconstruct the measured TDOS.
Instead, the same broad collective relaxation structure is recovered
independently.
The large-scale spectral organization observed in the main analysis
is therefore stable against the change in numerical precision.

The infrared comparison is shown more clearly in
Fig.~\ref{fig:fp64_fullblock_tdos}(b), where the spectra are restricted
to
\begin{equation}
\lambda\leq0.05
\end{equation}
and evaluated using common logarithmic bins.
Over the well-populated infrared regime, the FP32 and FP64 spectra
again recover closely related spectral densities and the same overall
low-$\lambda$ organization.
Differences become more visible only toward the extreme infrared,
where the number of sampled modes becomes small and individual
histogram bins consequently exhibit larger statistical fluctuations.

The FP64 calculation additionally allows the numerical continuation
of the spectrum to be examined at relaxation rates substantially
smaller than those used in the main-text scaling analysis.
The measured slow-mode tail extends into the vicinity of
\begin{equation}
\lambda
\sim
10^{-8},
\end{equation}
with the smallest populated bins reaching this extreme-infrared
regime.
The observation of modes at such small relaxation rates should not be
interpreted as assigning eight-decimal-digit physical precision to the
original pretrained parameters.
Rather, it demonstrates that when the subsequent Jacobian calculation
is performed with increased numerical precision, the collective
spectrum does not terminate or qualitatively change at the
relaxation-rate scales emphasized in the FP32 analysis.

It is useful to distinguish the \emph{extreme infrared extent} of the
measured spectrum from the \emph{resolved infrared scaling window}
used to determine the exponent in the main text.
The cumulative scaling analysis was deliberately performed over the
fixed interval
\begin{equation}
10^{-4}
\leq
\lambda
\leq
5\times10^{-2},
\label{eq:fp64_scaling_window}
\end{equation}
where all retained sequence lengths provide sufficiently dense mode
statistics for a common fit.
The much smaller relaxation rates visible in
Fig.~\ref{fig:fp64_fullblock_tdos}(b) are therefore not used to
determine the reported infrared exponent.

These two spectral regimes test different aspects of the result.
The common fitting interval establishes the reproducible collective
scaling
\begin{equation}
\rho_{\rm block}(\lambda)
\sim
\lambda^{\beta_{\rm IR}},
\qquad
\beta_{\rm IR}\simeq-0.17,
\end{equation}
over a statistically well-resolved region.
The extreme-infrared measurement instead tests whether the collective
relaxation hierarchy continues toward substantially slower modes when
the numerical precision of the Jacobian analysis is increased.
The latter therefore provides a robustness test rather than an
extrapolation of the fitted power law to
$\lambda\rightarrow0$.

This distinction is also important because the increasing TDOS toward
the extreme infrared does not imply that this region contains the
largest absolute number of modes.
The TDOS is a density with respect to relaxation rate.
Under logarithmic visualization, a small number of modes occupying a
very narrow interval in $\lambda$ can therefore produce a large local
density even when the cumulative population of such modes remains
small.
The stronger fluctuations observed in the lowest-$\lambda$ bins are
consistent with this sparse-mode character and should not be confused
with the densely resolved continuum used for the scaling analysis.

The FP32--FP64 comparison nevertheless shows that this statistical
distinction does not alter the central result.
Over the resolved collective regime, the spectra obtained at the two
numerical precisions reproduce essentially the same TDOS morphology.
The weakly varying slow-mode sector, its infrared enhancement, and the
subsequent crossover toward rapidly relaxing modes all persist under
the FP64 calculation.
At still smaller relaxation rates, the FP64 analysis extends the
observable tail toward $\lambda\sim10^{-8}$ without producing a
qualitatively different spectral organization.

Taken together, these results provide an independent numerical
validation of the full-block collective spectrum reported in the main
text.
The agreement between FP32 and FP64 calculations demonstrates that
the observed infrared organization is not generated simply by the
numerical precision of the Jacobian analysis, while the extension of
the FP64 spectrum toward extremely small relaxation rates confirms
the persistence of a hierarchy of very slow collective modes.
At the same time, the infrared exponent reported in the main text is
determined only from the substantially more conservative,
well-populated common scaling window and therefore does not depend on
the numerical interpretation of the extreme-infrared tail.

\clearpage
\appendix

\renewcommand{\thefigure}{D\arabic{figure}}
\renewcommand{\theequation}{D\arabic{equation}}

\setcounter{figure}{0}
\setcounter{equation}{0}

\section{Appendix D: Memory Kernel from the Full-Block Relaxation Spectrum}
\label{app:memory_kernel}

The full-block Jacobian measurements presented in the main text
characterize the collective relaxation spectrum of the complete Mamba
block.
Here we examine the same organization directly in the time domain by
constructing the memory kernel from the measured relaxation modes.
This provides a complementary representation of the infrared TDOS and
makes explicit how the observed accumulation of slow collective modes
generates long-time temporal persistence.

For a set of stable full-block Jacobian modes with relaxation rates
$\{\lambda_\alpha\}$, the normalized spectral memory kernel is defined
as
\begin{equation}
K(t)
=
\frac{1}{N}
\sum_{\alpha=1}^{N}
e^{-\lambda_\alpha t},
\label{eq:appB_discrete_kernel}
\end{equation}
where $N$ is the number of modes included in the corresponding
spectral sector.
In the continuum limit, this becomes
\begin{equation}
K(t)
=
\int d\lambda\,
\rho(\lambda)e^{-\lambda t},
\label{eq:appB_continuum_kernel}
\end{equation}
where $\rho(\lambda)$ is the normalized time-scale density of states.
The memory kernel is therefore the Laplace transform of the measured
relaxation spectrum.

Equation~(\ref{eq:appB_continuum_kernel}) immediately shows why the
infrared sector controls long-time dynamics.
At short times, relaxation modes over a broad spectral range
contribute to the response.
As time increases, modes with large $\lambda$ are exponentially
suppressed by the factor $e^{-\lambda t}$, and the remaining response
becomes progressively dominated by modes with small relaxation rates.
The long-time behavior of $K(t)$ therefore provides a direct
time-domain representation of the low-$\lambda$ organization observed
in the full-block TDOS.

\begin{figure}[t]
\centering
\includegraphics[scale=0.59, trim= 0.3cm 0.5cm 0cm 0cm]{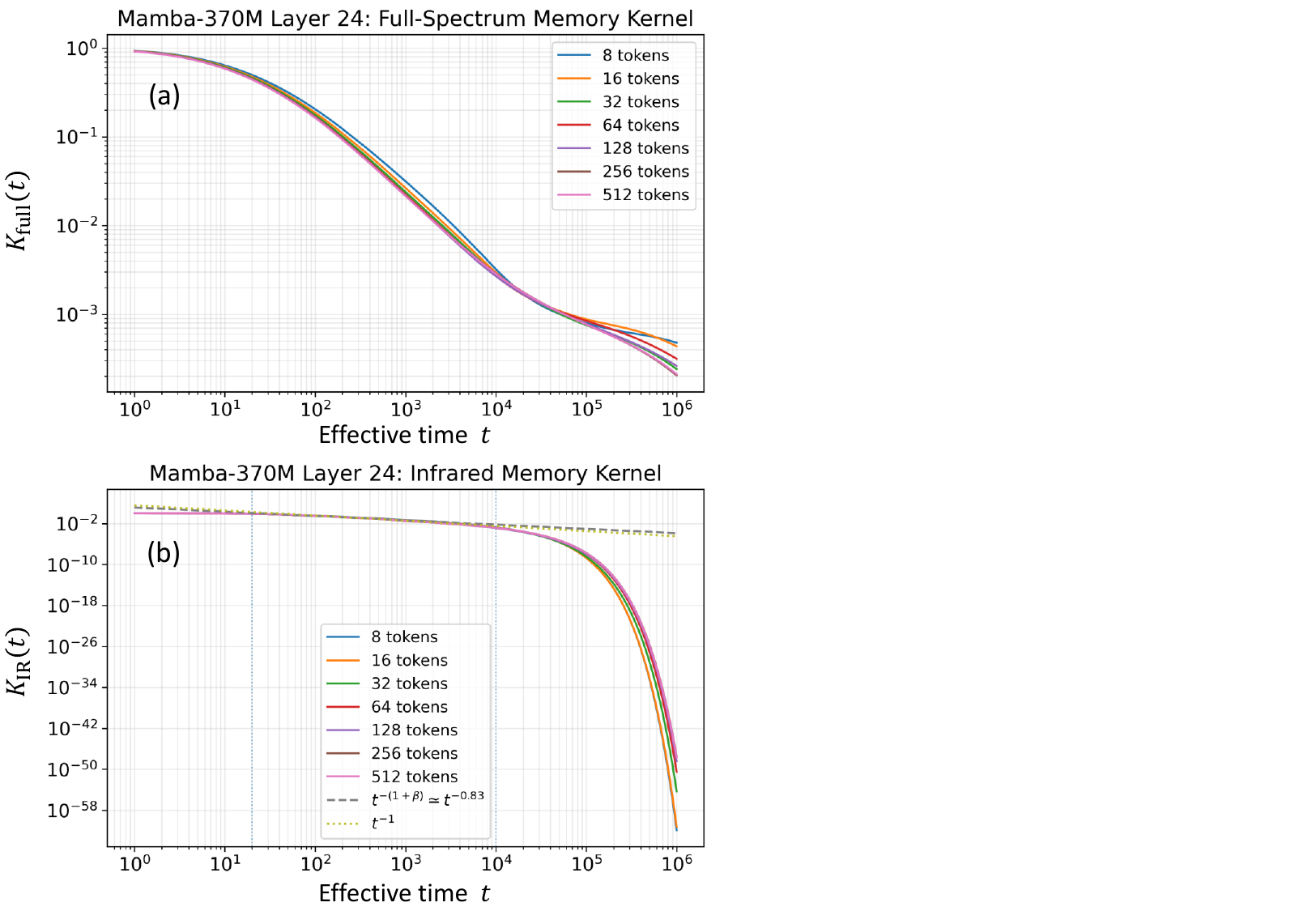}
\caption{
Memory kernels constructed from the full-block Jacobian relaxation
spectrum of Mamba-370M at layer~24 for retained sequence lengths
$L=8$--$512$.
(a) Full-spectrum memory kernel
$K_{\rm full}(t)=N^{-1}\sum_{\alpha}e^{-\lambda_\alpha t}$,
evaluated using all measured stable relaxation modes.
Despite the substantial increase in the number of resolved modes with
sequence length, the kernels exhibit closely overlapping short- and
intermediate-time behavior, while differences become increasingly
visible at long times as larger sequence subspaces resolve additional
deep-infrared modes.
The resulting extension of the long-time tail directly reflects the
progressively resolved low-relaxation-rate spectral weight.
(b) Infrared memory kernel $K_{\rm IR}(t)$ constructed using only modes
within the common scaling interval
$10^{-4}\leq\lambda\leq5\times10^{-2}$ employed for the cumulative
TDOS analysis.
For a power-law TDOS $\rho(\lambda)\sim\lambda^\beta$, the corresponding
intermediate-time kernel obeys
$K_{\rm IR}(t)\sim t^{-(1+\beta)}$ over
$\lambda_{\max}^{-1}\ll t\ll\lambda_{\min}^{-1}$.
The vertical dotted lines mark the inverse spectral boundaries,
$t_{\rm UV}\simeq20$ and $t_{\rm IR}=10^4$.
The dashed $t^{-0.83}$ reference corresponds to the measured
long-sequence exponent $\beta_{\rm IR}\simeq-0.17$, while the dotted
$t^{-1}$ line denotes the exactly flat-TDOS limit $\beta=0$.
These reference lines are scaling guides rather than fits to the
kernel.
Beyond $t_{\rm IR}$, the finite lower cutoff of the selected infrared
window produces the expected rapid decay.
Together, the two panels show how the measured collective relaxation
spectrum is transformed into extended temporal memory and distinguish
the long-time contribution of the complete deep-infrared spectrum from
the scaling behavior associated with the fixed resolved infrared
window.
}
\label{fig:mamba_memory_kernels}
\end{figure}

Figure~\ref{fig:mamba_memory_kernels}(a) shows the memory kernel
constructed from the complete measured stable-mode spectrum at
layer~24 for retained sequence lengths from 8 to 512 tokens.
Despite the substantial increase in the number of resolved modes with
sequence length, the kernels exhibit closely related short- and
intermediate-time behavior.
Differences become increasingly visible at long times, where the
kernel is most sensitive to the progressively resolved deep-infrared
modes.
The persistence of the full-spectrum kernel to long effective times
therefore reflects the extension of the measured collective spectrum
toward increasingly small relaxation rates.

The connection between the infrared TDOS and the temporal scaling of
the memory kernel can be obtained analytically.
Suppose that over a finite infrared interval the TDOS is described by
\begin{equation}
\rho(\lambda)
=
C\lambda^\beta,
\qquad
\lambda_{\min}
\leq
\lambda
\leq
\lambda_{\max}.
\label{eq:appB_powerlaw_tdos}
\end{equation}
The corresponding infrared contribution to the memory kernel is
\begin{equation}
K_{\rm IR}(t)
=
C
\int_{\lambda_{\min}}^{\lambda_{\max}}
d\lambda\,
\lambda^\beta e^{-\lambda t}.
\label{eq:appB_ir_kernel}
\end{equation}

Introducing the dimensionless variable
\begin{equation}
u=\lambda t,
\qquad
d\lambda=\frac{du}{t},
\end{equation}
gives
\begin{align}
K_{\rm IR}(t)
&=
C
\int_{\lambda_{\min}t}^{\lambda_{\max}t}
\frac{du}{t}
\left(
\frac{u}{t}
\right)^\beta
e^{-u}
\nonumber\\
&=
C t^{-(1+\beta)}
\int_{\lambda_{\min}t}^{\lambda_{\max}t}
du\,
u^\beta e^{-u}.
\label{eq:appB_scaled_kernel}
\end{align}
Equivalently,
\begin{equation}
K_{\rm IR}(t)
=
C t^{-(1+\beta)}
\left[
\gamma(1+\beta,\lambda_{\max}t)
-
\gamma(1+\beta,\lambda_{\min}t)
\right],
\label{eq:appB_gamma_kernel}
\end{equation}
where $\gamma(a,x)$ is the lower incomplete gamma function.

This expression naturally produces three temporal regimes.
At short times,
\begin{equation}
t\ll\lambda_{\max}^{-1},
\end{equation}
all modes in the selected spectral interval satisfy $\lambda t\ll1$.
Using
\begin{equation}
e^{-\lambda t}
\simeq
1-\lambda t+\cdots,
\end{equation}
the leading contribution becomes
\begin{equation}
K_{\rm IR}(t)
\simeq
\frac{C}{1+\beta}
\left(
\lambda_{\max}^{1+\beta}
-
\lambda_{\min}^{1+\beta}
\right).
\label{eq:appB_short_kernel}
\end{equation}
The kernel is therefore approximately constant at sufficiently short
times.

At intermediate times satisfying
\begin{equation}
\lambda_{\max}^{-1}
\ll
t
\ll
\lambda_{\min}^{-1},
\label{eq:appB_scaling_window_time}
\end{equation}
the scaled limits obey
\begin{equation}
\lambda_{\min}t\ll1,
\qquad
\lambda_{\max}t\gg1.
\end{equation}
Equation~(\ref{eq:appB_scaled_kernel}) then reduces to
\begin{align}
K_{\rm IR}(t)
&\simeq
C t^{-(1+\beta)}
\int_0^\infty
du\,
u^\beta e^{-u}
\nonumber\\
&=
C\Gamma(1+\beta)
t^{-(1+\beta)}.
\label{eq:appB_powerlaw_kernel}
\end{align}
Thus, within the temporal regime associated with the resolved
power-law spectral window,
\begin{equation}
\rho(\lambda)\sim\lambda^\beta
\qquad\Longrightarrow\qquad
K(t)\sim t^{-(1+\beta)}.
\label{eq:appB_tdos_kernel_relation}
\end{equation}

For an exactly flat TDOS,
\begin{equation}
\beta=0,
\end{equation}
this relation gives the marginal memory form
\begin{equation}
K(t)\sim t^{-1}.
\label{eq:appB_flat_kernel}
\end{equation}
For the Mamba full-block spectrum measured in the main text, the
long-sequence cumulative analysis gives approximately
\begin{equation}
\beta_{\rm IR}\simeq-0.17.
\end{equation}
The corresponding temporal scaling is therefore
\begin{equation}
K_{\rm IR}(t)
\sim
t^{-0.83}.
\label{eq:appB_mamba_kernel}
\end{equation}
The measured Mamba collective dynamics thus lies close to the
marginal $1/t$ regime, with a weak enhancement of long-time
persistence.

At still longer times,
\begin{equation}
t\gg\lambda_{\min}^{-1},
\end{equation}
even the slowest modes contained in the selected spectral interval are
suppressed.
The leading asymptotic contribution then has the form
\begin{equation}
K_{\rm IR}(t)
\propto
\frac{\lambda_{\min}^{\beta}}{t}
e^{-\lambda_{\min}t},
\label{eq:appB_long_kernel}
\end{equation}
up to subleading corrections.
A finite power-law relaxation spectrum therefore produces the
characteristic temporal sequence
\begin{equation}
\text{short-time plateau}
\;\rightarrow\;
t^{-(1+\beta)}
\;\rightarrow\;
\text{infrared-cutoff decay}.
\label{eq:appB_three_regimes}
\end{equation}

To compare this prediction directly with the spectral measurements,
we construct an infrared memory kernel using the same fixed interval
employed for the cumulative TDOS analysis in the main text,
\begin{equation}
10^{-4}
\leq
\lambda
\leq
5\times10^{-2}.
\label{eq:appB_fixed_ir_window}
\end{equation}
The two characteristic temporal scales associated with these spectral
boundaries are
\begin{equation}
t_{\rm UV}
=
\lambda_{\max}^{-1}
=
\frac{1}{5\times10^{-2}}
\simeq20,
\label{eq:appB_tuv}
\end{equation}
and
\begin{equation}
t_{\rm IR}
=
\lambda_{\min}^{-1}
=
\frac{1}{10^{-4}}
=
10^4.
\label{eq:appB_tir}
\end{equation}
The power-law regime implied by the measured infrared spectrum is
therefore expected primarily over
\begin{equation}
20
\lesssim
t
\lesssim
10^4.
\label{eq:appB_expected_time_window}
\end{equation}

Figure~\ref{fig:mamba_memory_kernels}(b) shows the corresponding
infrared memory kernel calculated directly from the measured modes in
this fixed spectral interval.
The kernel first exhibits a weakly varying short-time regime and then
enters an extended slowly decaying region between the inverse spectral
boundaries.
The measured long-sequence exponent
$\beta_{\rm IR}\simeq-0.17$ predicts
$K_{\rm IR}(t)\sim t^{-0.83}$ in this region, close to the
$t^{-1}$ behavior associated with an exactly flat TDOS.
The two reference power laws in Fig.~\ref{fig:mamba_memory_kernels}(b)
are therefore shown as scaling guides rather than independent fits to
the memory kernel.

For $t\gtrsim10^4$, the kernel calculated from the restricted
infrared interval falls rapidly.
This decay should not be interpreted as the disappearance of memory
from the complete Mamba spectrum.
It follows directly from imposing the lower boundary
$\lambda_{\min}=10^{-4}$ in the restricted construction.
The full spectrum contains relaxation modes extending substantially
below this value, and these additional modes continue to contribute at
longer effective times, as directly demonstrated by
Fig.~\ref{fig:mamba_memory_kernels}(a).

The two panels therefore probe complementary aspects of the same
collective relaxation spectrum.
Figure~\ref{fig:mamba_memory_kernels}(a) retains the complete measured
spectrum and demonstrates the long temporal reach generated by the
deepest resolved relaxation modes.
Figure~\ref{fig:mamba_memory_kernels}(b) isolates the common infrared
scaling window and directly tests the temporal consequence of the TDOS
exponent extracted in the main text.
The distinction is important because the cutoff of the restricted
kernel is determined by the chosen lower spectral boundary, whereas
the full-spectrum kernel continues to receive contributions from modes
at substantially smaller relaxation rates.

The time-domain analysis therefore provides an independent
representation of the infrared organization identified from the
full-block Jacobian spectrum.
The cumulative TDOS establishes a nearly flat but weakly
infrared-enhanced regime,
$\rho(\lambda)\sim\lambda^{-0.17}$, and its Laplace transform produces
the corresponding slowly decaying response
$K(t)\sim t^{-0.83}$ over the associated temporal window.
The proximity of this behavior to the marginal $1/t$ form connects
the measured spectral accumulation of slow collective modes directly
to extended temporal persistence.

Importantly, the memory kernel considered here is constructed from the
collective relaxation spectrum of the complete Mamba block.
It should therefore be distinguished from the microscopic SSM
convolution kernel $K_{\rm SSM}(t)=Ce^{At}B$ discussed in Sec.~II.
The latter propagates the history of the external input through the
explicit state-space dynamics, whereas the present kernel characterizes
the temporal organization implied by the collective full-block
relaxation spectrum.
This distinction preserves the microscopic-to-macroscopic hierarchy
central to the analysis: the explicit SSM supplies a microscopic
state-space memory mechanism, while the full-block Jacobian reveals
the emergent collective relaxation structure whose infrared spectrum
generates the long-time kernel examined here.


\begin{thebibliography}{99}


\bibitem{1}
B. G. Chae,
``Infrared organization and critical cognitive field formation in
Transformer dynamics,''
arXiv preprint arXiv:2607.10923v4 (2026).

\bibitem{2}
A. Vaswani, N. Shazeer, N. Parmar, J. Uszkoreit, L. Jones,
A. N. Gomez, L. Kaiser, and I. Polosukhin,
``Attention is all you need,''
Advances in Neural Information Processing Systems \textbf{30} (2017).

\bibitem{3} A. Radford, K. Narasimhan, T. Salimans, and I. Sutskever, 
``Improving language understanding by generative pre-training,'' OpenAI (2018).

\bibitem{4} T. B. Brown, B. Mann, N. Ryder, M. Subbiah, J. Kaplan et al.,
``Language models are few-shot learners," In Advances in Neural Information Processing Systems, 1877-1901 (2020).

\bibitem{5} J. Hoffmann, S. Borgeaud, A. Mensch, E. Buchatskaya, T. Cai, E. Rutherford et al., ``Training compute-optimal large language models,'' In Proceedings of the 36th International Conference on Neural Information Processing Systems. 30016-30030 (2022).

\bibitem{6}J. Kaplan, S. McCandlish, T. Henighan, T. B. Brown, B. Chess, R. Child, S. Gray, A. Radford, J. Wu, and D. Amodei, ``Scaling laws for neural language models,'' arXiv preprint arXiv.2001.08361 (2020).


\bibitem{7}
A. Gu, T. Dao, S. Ermon, A. Rudra, and C. R\'e,
``HiPPO: Recurrent memory with optimal polynomial projections,''
Advances in Neural Information Processing Systems \textbf{33}
(2020).


\bibitem{8}
A. Gu, K. Goel, and C. R\'e,
``Efficiently modeling long sequences with structured state spaces,''
International Conference on Learning Representations (2022).


\bibitem{9}
A. Gu, A. Gupta, K. Goel, and C. R\'e,
``On the parameterization and initialization of diagonal state space
models,''
Advances in Neural Information Processing Systems \textbf{35},
35971--35983 (2022).


\bibitem{10}
J. T. H. Smith, A. Warrington, and S. W. Linderman,
``Simplified state space layers for sequence modeling,''
International Conference on Learning Representations (2023).


\bibitem{11}
D. Y. Fu, T. Dao, K. K. Saab, A. W. Thomas, A. Rudra, and C. R\'e,
``Hungry hungry hippos: Towards language modeling with state space
models,''
International Conference on Learning Representations (2023).


\bibitem{12}
A. Orvieto, S. L. Smith, A. Gu, A. Fernando, C. G\"ul\c{c}ehre,
R. Pascanu, and S. De,
``Resurrecting recurrent neural networks for long sequences,''
Proceedings of the 40th International Conference on Machine Learning,
PMLR \textbf{202} (2023).


\bibitem{13}
M. Poli, S. Massaroli, E. Nguyen, D. Y. Fu, T. Dao, S. Baccus,
Y. Bengio, S. Ermon, and C. R\'e,
``Hyena hierarchy: Towards larger convolutional language models,''
Proceedings of the 40th International Conference on Machine Learning,
PMLR \textbf{202}, 28043--28078 (2023).



\bibitem{14}
A. Gu and T. Dao,
``Mamba: Linear-time sequence modeling with selective state spaces,''
arXiv preprint arXiv:2312.00752 (2023).


\bibitem{15}
N. M. Cirone, A. Orvieto, B. Walker, C. Salvi, and T. Lyons,
``Theoretical foundations of deep selective state-space models,''
Advances in Neural Information Processing Systems \textbf{37}
(2024).

\bibitem{16}
T. Dao and A. Gu,
``Transformers are SSMs: Generalized models and efficient algorithms
through structured state space duality,''
Proceedings of the 41st International Conference on Machine Learning,
PMLR \textbf{235}, 10041--10071 (2024).



\bibitem{17}
B. G. Chae,
``Cognitive field theory: Memory-dressed collective dynamics of
intelligence,''
arXiv preprint arXiv:2601.10221v8 (2026).



\bibitem{18}
S. Biderman, H. Schoelkopf, Q. Anthony, H. Bradley, K. O'Brien,
E. Hallahan, M. A. Khan et al.,
``Pythia: A suite for analyzing large language models across training
and scaling,''
Proceedings of the 40th International Conference on Machine Learning,
PMLR \textbf{202}, 2397--2430 (2023).


\end{thebibliography}
\end{document}